\pdfoutput=1

\documentclass[11pt]{article}

\usepackage[preprint]{acl}

\usepackage{times}
\usepackage{latexsym}

\usepackage{amssymb}

\usepackage[T1]{fontenc}

\usepackage[utf8]{inputenc}

\usepackage{microtype}

\usepackage{inconsolata}

\usepackage{graphicx}

\usepackage{multirow}

\usepackage{listings,mdframed}
\usepackage[most]{tcolorbox}

\newcommand{\ctelm}{\texttt{ctELM}}

\newcommand{\embToAbst}{\texttt{emb2abs}}
\newcommand{\embToSect}{\texttt{emb2sec}}
\newcommand{\embToPls}{\texttt{emb2pls}}
\newcommand{\embToCommonalities}{\texttt{emb2com}}
\newcommand{\embToDifferences}{\texttt{emb2dif}}

\title{ctELM: Decoding and Manipulating Embeddings of Clinical Trials with Embedding Language Models}

\author{Brian Ondov*, Chia-Hsuan Chang*, Yujia Zhou, Mauro Giuffrè \and Hua Xu \\
  Department of Biomedical Informatics \& Data Science \\
  Yale School of Medicine \\
  New Haven, CT, USA \\
  \texttt{\{brian.ondov,chia-hsuan.chang,yujia.zhou,mauro.giuffre,hua.xu\}@yale.edu}}

\begin{document}
\maketitle
\def\thefootnote{*}\footnotetext{These authors contributed equally to this work}\def\thefootnote{\arabic{footnote}}
\begin{abstract}
Text embeddings have become an essential part of a variety of language applications.
However, methods for interpreting, exploring and reversing embedding spaces are limited, reducing transparency and precluding potentially valuable generative use cases.
In this work, we align Large Language Models to embeddings of clinical trials using the recently reported Embedding Language Model (ELM) method.
We develop an open-source, domain-agnostic ELM architecture and training framework, design training tasks for clinical trials, and introduce an expert-validated synthetic dataset.
We then train a series of ELMs exploring the impact of tasks and training regimes. Our final model, ctELM, can accurately describe and compare unseen clinical trials from embeddings alone and produce plausible clinical trials from novel vectors. We further show that generated trial abstracts are responsive to moving embeddings along concept vectors for age and sex of study subjects. Our public ELM implementation and experimental results will aid the alignment of Large Language Models to embedding spaces in the biomedical domain and beyond.
\end{abstract}

\section{Introduction}

Text embeddings map variable-length text documents to fixed-length vector spaces, capturing rich semantic information. These embeddings have become ubiquitous in Natural Language Processing, owed to the utility of embedding spaces for similarity scoring, classification, and other applications~\cite{muennighoff2023mteb}. However, embedding of text is typically a one-way process, and the resulting embeddings are treated as `black boxes,' useful for applications but uninterpretable and irreversible. In addition to making embedding spaces more interpretable, reversing these spaces can aid in producing more creative content~\cite{yeh2025exploring, zhang2025mapexplorer}. Yet, existing methods for inverting embeddings are extremely limited in length do not invert arbitrary vectors well, and cannot perform more advanced reasoning over one or more vectors.~\cite{song2020information,morris-etal-2023-text, tennenholtzDemystifyingEmbeddingSpaces2024}.

A promising potential solution is training Embedding Language Models (ELMs) to allow interaction with embeddings via natural language~\cite{tennenholtzDemystifyingEmbeddingSpaces2024}. ELMs extend language models by adding an adapter layer that aligns an embedding space of interest to the model's own token embedding space, allowing prompts to contain mixtures of tokens and complete text embeddings.
In addition to enabling operations over arbitrary vectors, ELMs have been shown to more faithfully represent interpolated embedding vectors than text-only LLMs prompted to combine original text inputs.
Still, ELMs have only been reported for the narrow domain of film reviews, and for proprietary base language models, with no open-source codebase for implementing or training them. Questions also remain around optimal training procedures. 

In this work, we seek to advance methods for making embeddings and embedding spaces more transparent. We build on the work of~\citet{tennenholtzDemystifyingEmbeddingSpaces2024} by creating an open-source ELM architecture and training framework and by exploring the viability of ELMs in the biomedical domain. We use our new implementation to align an ELM to embeddings of clinical trials, designing domain-specific training tasks and constructing an expert-validated dataset. In extensive experiments, we demonstrate that our model, \ctelm{}, can reconstruct abstracts more reliably than Vec2Text and can perform additional tasks requiring reasoning over multiple embeddings. We show \ctelm{} can produce plausible, hypothetical clinical trials from novel embedding vectors obtained by interpolating or perturbing embeddings derived from text sources. Further, we show that the generated abstracts are responsive to clinically meaningful directions identified in the embedding space using Concept Activation Vectors~\cite{kim2018interpretability}, namely those representing the sex and age of trial subjects. Finally, we advance general knowledge of ELMs by performing extensive ablations showing the effects of tasks, training regimes, embedding models, and generation parameters.
The main contributions of this work are:
(1) the first open-source ELM architecture and training framework; (2) an expert-validated dataset for training ELMs to interpret embeddings of clinical trials; (3) a trained ELM that can interpret embeddings of clinical trials; and (4) ablation studies adding to knowledge of optimal ELM training.

\begin{figure*}[t]
  \includegraphics[width=\textwidth]{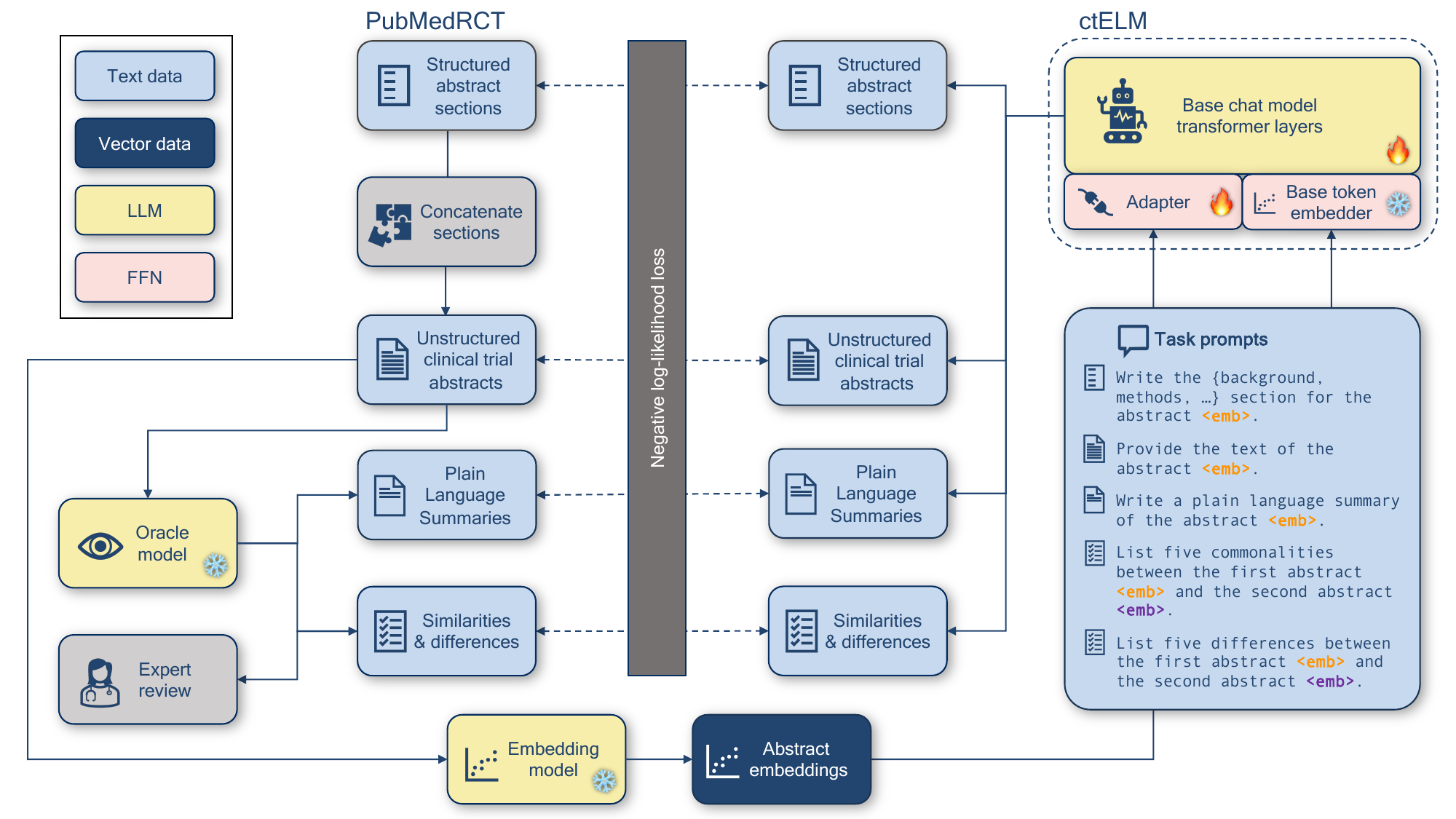}
  \caption{The data generation and training pipeline for ctELM.}
  \label{fig:data_training_pipeline_ctELM}
\end{figure*}

\section{Background}

\subsection{Text Embeddings}

Early, ``shallow'' embeddings operated on individual tokens and were learned using neural networks with single hidden layers or matrix factorization methods~\cite{mikolov2013distributed, pennington2014glove}. Following the introduction of pretrained transformers~\cite{vaswani2017attention, radford2019language, devlin2019bert}, sentence-level, and subsequently document-level text embeddings became viable using contrastive learning on pooled contextual embeddings for individual tokens~\cite{reimers2019sentence,gao2021simcse,behnamghader2024llm2vec,xiao_etal_2024_bge}.

\subsection{Inversion as Vulnerability}

One line of research on embeddings assumes an attacker trying to access private information that would have been thought to be inherently secure due to the `black box' nature of embeddings~\cite{song2020information}. This line gave rise to GEIA (Generative Embedding Inversion Attack)~\cite{li2023sentence}. GEIA projects a vector embedding to the token embedding layer in place of the first token of input to a decoder-only transformer-based language model, in this case using the GPT-2 architecture~\cite{radford2019language}. The model is then trained from random weights using teacher forcing to recover the original sentence one token at a time.

Building on this work, Vec2Text~\cite{morris-etal-2023-text} fine-tunes encoder-decoder transformer language models to become an `inverter' and a `corrector,' in this case both based on pretrained T5~\cite{raffel2020exploring}. Given an embedding, the inverter is trained to make an initial hypothesis of the original text, and the corrector is trained to move the hypothesis text closer to the target embedding by making discrete updates, based on both the target embedding and the embedding of the current hypothesis. Given enough iterative updates, this method can accurately recover short text sequences from embeddings alone. The introduction of Vec2Text has spurred subsequent research on defending against embedding inversion attacks~\cite{zhuang2024understanding}. A reproduction study of Vec2Text by~\citet{seputis2025rethinking} also confirmed the major findings of~\citet{morris-etal-2023-text}. Aside from fixed-length semantic embeddings,~\citet{kugler2024invbert} showed that text can also be recovered from the token-level contextual embeddings of BERT~\cite{devlin2019bert}.


\subsection{Vector-Controlled Generation}

Another line of work seeks to understand and reverse embedding spaces in order to use directions in those spaces to control content. The beginnings of this paradigm can be seen in \citealp{bolukbasi2016man}, in which a gender axis is identified in a shallow word embedding space, and this axis neutralized in word embeddings that are undesirably gendered. More recently,~\citet{tennenholtzDemystifyingEmbeddingSpaces2024} introduce Embedding Language Models, partly with the aim of exploring embedding spaces to generate more novel text content. This work explored moving embeddings of film plots with reviews along axes (representing attributes such as comedy or drama) identified in the embedding space using Concept Activation Vectors (CAVs)~\cite{kim2018interpretability}. CAVs are essentially vectors orthogonal to the decision plane of a linear classifier for a concept of interest and were originally developed for explaining predictions of vision models based on internal activations.
Steerability in LLMs has also been explored using Concept Bottlenecks~\cite{DBLP:conf/iclr/SunOUW25}. However, these require labels during LLM training, making them less flexible than aligning to an embedding space, and the training process significantly degrades language modeling ability.

\section{Methods}

\subsection{Preliminaries}

Let the embedding model $E_{emb}$ denote a mapping from a sequence of language tokens to an embedding space, formally expressed as: $E_{emb}: \mathcal{X} \mapsto \mathcal{Z}_{emb}$, where the length of the token sequence $\mathcal{X}$ is bounded by the context length limitation inherent to $E_{emb}$.
The base chat model, represented by $\mathcal{M}$, is a text-only, instruction-tuned large language model (LLM). Formally, this model translates one sequence of language tokens into another: $\mathcal{M}: \mathcal{X} \mapsto \mathcal{X}$. The chat model comprises two primary components, namely the token embedding layer $E_{base}$ and the transformer layers $M_{base}$. The token embedding layer $E_{base}$ maps input tokens to a token embedding space: $E_{base}: \mathcal{X} \mapsto \mathcal{Z}_{base}$, where each token $x \in \mathcal{X}$ is encoded into a token embedding vector $z \in \mathcal{Z}_{base}$. The transformer layers $M_{base}$ subsequently transforms these token embeddings into output token sequences: $M_{base}: \mathcal{Z}_{base} \mapsto \mathcal{X}$.

\subsection{Architecture}

Our objective is to extend the base model $\mathcal{M}$ to a target model ($\mathcal{M}_{tgt}$) capable of processing and interpreting embeddings produced by the external embedding model $E_{emb}$, alongside standard language tokens. Specifically, $\mathcal{M}_{tgt}$ operates on a combination of tokens and embeddings to produce text output: $\mathcal{M}_{tgt}: (\mathcal{X},\mathcal{Z}_{emb}) \mapsto \mathcal{X}$. 
To accomplish this, we adapt the approach proposed by~\citet{tennenholtzDemystifyingEmbeddingSpaces2024}. We introduce an adapter module $\mathcal{A}$ to align the embedding spaces of $\mathcal{Z}_{emb}$ and $\mathcal{Z}_{base}$. Consequently, the target model is defined as: $\mathcal{M}_{tgt} = (\mathcal{A},E_{base},M_{base})$. The adapter $\mathcal{A}$ is a two-layer multilayer perceptron (MLP):

\begin{equation}
   W_1( \sigma(W_0Z_{emb}+b_0))+b_1,
\end{equation}

\noindent where $W_0, b_0, W_1, b_1$ are learnable weights and $\sigma$ is a non-linear activation function. The adapter ensures that the embeddings $Z_{emb}$ produced by the external embedding model $E_{emb}$ are projected into the embedding space $\mathcal{Z}_{base}$, thereby enabling $M_{base}$ to generate output texts by effectively leveraging both token-derived contextual information ($\mathcal{Z}_{base}$) and the semantic content encoded within $\mathcal{Z}_{emb}$. In this sense, the ELM architecture has similarities with Vision Language Models~\cite{zhang2024vision}, which must also use adapters to align language models to a dense vector space containing visual information.

\subsection{Data \& Task Preparation}

We use PubMed 200K RCT~\cite{dernoncourt2017pubmed}, which contains 190,654, 2,500, and 2,500 abstracts for training, validation, and testing sets, respectively. This dataset was created to aid in classifying structured abstract sections; here we will use it to generate those sections, and as a clean collection of randomized controlled trials.  Since each abstract is structured and separated by section, we concatenate all sections into an unstructured clinical trial abstract. A selected embedding model $E_{emb}$ is then employed to generate embedding $z \in \mathcal{Z}_{emb}$ for each abstract.

ELMs are aligned to embedding spaces by performing various tasks that would require detailed knowledge of the content of the embeddings. To train \ctelm{} using the $\mathcal{M}_{tgt}$ model architecture, we prepare five diverse tasks relevant to clinical trial abstracts, as illustrated in Fig.~\ref{fig:data_training_pipeline_ctELM}. A training instance is formulated as the input $p$ and the output $o$. The input $p$ is constructed as a prompt combining text tokens $\mathcal{X}$ (i.e., the task instruction) and abstract embedding $z \in Z_{emb}$. The output $o$ is the targeted text. Both input and output vary depending on the task. The following are the details of five different tasks (see Table~\ref{tab:data-statistic} for statistics):

\vspace{.5em}

\noindent \textbf{\embToAbst}: Decode an abstract embedding back to an abstract; $p$ is ``Provide the text of the abstract $z$'', $o$ is the original abstract text.
\vspace{.5em}

\noindent \textbf{\embToSect}: Decode an abstract embedding to a specific section. The input $p$ is asks to generate texts for a section from an abstract embedding $z$: ``Write the \{background, objective, method, result, or conclusion\} section for the abstract $z$''; $o$ is the corresponding section text. To balance the size of training samples across tasks, for each abstract we randomly sample a section from the abstract. 
\vspace{.5em}

\noindent \textbf{\embToPls}: Generates a plain language summary from an abstract embedding $z$ with input prompt $p$: ``Write a plain language summary of the abstract $z$''; $o$ is the plain language summary generated by an oracle model (see Appendix~\ref{sec:data-preparation-pls}).
\vspace{.5em}

\noindent \textbf{\embToCommonalities}: Analyzes two abstract embeddings and lists five commonalities. The prompt $p$ is thus crafted as ``List five commonalities between the first abstract $z_i$ and the second abstract $z_j$'', where both $z_i, z_j \in \mathcal{Z}_{emb}$ are abstract embeddings; $o$ is the commonality analysis generated by the oracle model. We use topic modeling to select abstract pairs from the same topic and across different topics to ensure diversity (see Appendix~\ref{sec:data-preparation-com-dif} for details).
\vspace{.5em}

\noindent \textbf{\embToDifferences}: Lists five differences for two given abstract embeddings. The prompt is: ``List five differences between the first abstract $z_i$ and the second abstract $z_j$''; $o$ is the difference analysis generated by the oracle model.

\begin{table}[]
\caption{Statistics for five tasks.
}
\label{tab:data-statistic}
\resizebox{\columnwidth}{!}{%
\begin{tabular}{llll}
\hline
& \textbf{Training} & \textbf{Validation} & \textbf{Testing} \\ \hline \hline
Task1: \embToAbst          & 190,654  & 2,500      & 2,500   \\
Task2: \embToSect          & 190,654  & 2,500      & 2,500   \\
Task3: \embToPls           & 190,654  & 2,500      & 2,500   \\
Task4: \embToCommonalities & 241,794  & 3,126      & 3,126   \\
Task5: \embToDifferences   & 241,794  & 3,180      & 3,180   \\ \hline
\end{tabular}%
}
\end{table}

\subsection{Training Procedure}

Although we prompt \ctelm{} with both text instructions and abstract embedding(s), its training is similar to text-only language models in that it aims to predict the next word in a sequence. Therefore, we can optimize the \ctelm{} by minimizing the negative log-likelihood loss versus training outputs. We keep the token embedding layer $E_{base}$ frozen and optimize embedding adapter $\mathcal{A}$ (with its parameter $W_0,b_0,W_1,b_1$) and transformer layers $M_{base}$. For efficient fine-tuning, in practice we employ Low-Rank Adaptation (LoRA)~\cite{hu2022lora} to optimize $M_{base}$, thus tuning low-rank adapters rather than the original weights. We explore both \textbf{one-phase training} (1P), which only jointly optimizes $\mathcal{A}$ and $M_{base}$, and \textbf{two-phase training} (2P), which adds an initial step in which $M_{base}$  is frozen.

\section{Experiments}

\subsection{Implementation \& Settings}

For experiments, we set $\mathcal{M}$ to be \texttt{Llama-3.1-8B-Instruct} \cite{dubey2024llama}, $E_{emb}$ to be \texttt{BAAI/bge-large-en-v1.5}, and adopt \texttt{gpt-4o-mini} as the oracle model to generate the synthetic data for \embToPls, \embToCommonalities, and \embToDifferences. We load the base model via Huggingface's transformers package~\cite{wolf2020transformers} and extend it by introducing an adapter $\mathcal{A}$, which is a 2-layer MLP with 2,048 neurons in the hidden layer, 4,096 in the second layer (to match LLama 3.1's token embedding dimension), and ReLU activation between the layers. The training procedure is implemented using the HuggingFace \texttt{trl} and \texttt{peft} packages to ensure reproducibility. All the training hyperparameters are reported in Appendix~\ref{sec:training-hyperparameters}.

For inference, we set the temperature as 1 for all tasks. However, we find that repetition occurs when generating entire abstracts. Therefore, we experiment with imposing a repetition penalty using the methods of \citet{keskar2019ctrl}, with the authors' recommended penalty strength of 1.2 for \embToAbst{} task. For other tasks, we set the penalty at 1.0 (no penalization), as we do not observe this behavior for these tasks.

\subsection{Model Variants}

We explore various training configurations of \ctelm{} to assess the effects of data scale, task diversity, and training procedure. \textbf{Data scale}: We train \ctelm{} on two dataset sizes, 190K and 1.2M training instances. \textbf{Task diversity}: We construct three datasets to analyze the influence of task diversity while keeping the training size fixed at 190K: 

\vspace{.5em}
    \noindent \textbf{1-task}: Only \embToAbst{} training instances.

    \vspace{.5em}

    \noindent \textbf{3-task}: An equal number of instances from \embToAbst, \embToSect, and \embToPls.
    \vspace{.5em}

    \noindent \textbf{5-task}: Equally sampled instances from \embToAbst, \embToSect, \embToPls, \embToCommonalities, and \embToDifferences. 
\vspace{.5em}

For the 1.2M configuration, we interleave training instances from all five tasks, sampling until each instance from every task has been included at least once. \textbf{Training procedure}: We further examine the effect of training strategy by varying the number of training phases. Each model configuration is denoted as $x$P-$y$E, where $x$ indicates the number of distinct training phases and $y$ denotes the number of training epochs.


\subsection{Baselines}

As \ctelm{} is trained on clinical trial studies, direct comparison with the originally reported ELM of \citet{tennenholtzDemystifyingEmbeddingSpaces2024} is not feasible, since (1) it was trained on a movie review dataset and (2) the architecture is not publicly available for retraining. However, as the \texttt{emb2abs} task is direct inversion of an embedding, we can compare \texttt{ctELM}'s performance for this task against Vec2Text~\cite{morris-etal-2023-text}, to our knowledge the state-of-the art system for embedding inversion. We first compare against the published GTR-base model, trained on Wikipedia passages. Though not trained specifically on clinical trial abstracts, \citet{morris-etal-2023-text} demonstrate generalization of this model to various biomedical corpora. As a stronger baseline, we also use the published GTR-base weights as initializations for further training on our corpus. We use the model weights and implementation from the official repository.\footnote{\url{https://github.com/vec2text/vec2tex}} Additionally, published Vec2Text models were only trained with a maximum of 128 tokens and are known to perform poorly beyond this length~\cite{seputis2025rethinking}. As  our abstracts have a mean length of 304 tokens, we thus also experiment with using Vec2Text to invert embeddings of individual sections, then concatenating the results to reconstruct the complete abstract. In total, we test four configurations:

\vspace{.5em}

    \noindent \textbf{Vec2Text}: The published model directly inverts an abstract embedding into its corresponding text.
    \vspace{.5em}

    \noindent \textbf{Vec2Text-ft}: The published model is further trained on our training abstracts, with the maximum tokens increased from 128 to 512, then inverts an embedding of a full abstract.
    \vspace{.5em}

    \noindent \textbf{Vec2Text-sect}: Each section of an abstract is embedded and decoded with the published model, and the resulting outputs concatenated.
    \vspace{.5em}

    \noindent \textbf{Vec2Text-sect-ft}: The published model is further trained to invert embeddings of individual sections from PubMedRCT  abstracts, leaving the maximum tokens at 128. At test time, each section of an abstract is independently embedded and decoded, and the results concatenated, as in Vec2Text-sect.
\vspace{.5em}

To approximately match the number of training steps of our 1.2M \ctelm{} models, we train the fine-tuned models (Vec2Text-ft and Vec2Text-sect-ft) for 7 epochs on the 190K abstracts from the PubMedRCT training set.

\subsection{Metrics}

We adopt Semantic Consistency (SC)~\cite{tennenholtzDemystifyingEmbeddingSpaces2024} as our primary metric. SC measures the semantic closeness between two embeddings. Formally, assuming that the generated text is $\hat{o}$, we embed it and compare with the embedding of the target text $o$ (or for novel embeddings, compare with the novel embedding directly):

\begin{equation}
    SC(\hat{o},o)=\delta(E_{emb}(\hat{o}),E_{emb}(o)),
\end{equation}

\noindent where $\delta$ measures cosine similarity between embedding in the semantic space~$Z_{emb}$ produced by the $E_{emb}$. As a result, the higher SC suggests the model generates texts that better align with the semantical concept of the target text.

\subsection{Results}

\begin{table*}[htb!]
\caption{Semantic Consistency for 5 tasks on the test set, across training tasks and strategies. \ctelm{} is trained on either 190K or 1.2M data using the \texttt{bge-large-en-v1.5} embedding model. ($x$P-$y$E) represents $x$-phase training procedure is adopted for $y$ epochs. Mean values over the test set are shown with standard deviations. Best performance for each task is marked in bold.}
\label{tab:SC-table}
\resizebox{\textwidth}{!}{%
\begin{tabular}{llcccccc}
\hline
Data Size & Model & \multicolumn{1}{l}{\begin{tabular}[c]{@{}l@{}}\embToAbst\\ (penalty=1.2)\end{tabular}} & \multicolumn{1}{l}{\begin{tabular}[c]{@{}l@{}}\embToAbst\\ (penalty=1.0)\end{tabular}} & \multicolumn{1}{l}{\embToSect} & \multicolumn{1}{l}{\embToPls} & \multicolumn{1}{l}{\embToCommonalities} & \multicolumn{1}{l}{\embToDifferences} \\ \hline \hline
\multirow{4}{*}{Baseline} & Vec2Text & 0.70$\pm$0.08 & - & - & - & - & - \\
& Vec2Text-ft & 0.77$\pm$0.08 & - & - & - & - & - \\
& Vec2Text-sect & 0.82$\pm$0.07 & - & - & - & - & - \\
& Vec2Text-sect-ft & 0.82$\pm$0.06 & - & - & - & - & - \\ \hline
\multirow{5}{*}{\begin{tabular}[c]{@{}l@{}}ctELM on\\ 190K\end{tabular}} & 1-task (1P-1E) & 0.83$\pm$0.05 & 0.82$\pm$0.05 & - & - & - & - \\
 & 3-task (1P-1E) & 0.83$\pm$0.05 & 0.81$\pm$0.05 & 0.73$\pm$0.07 & 0.77$\pm$0.05 & - & - \\
 & 3-task (1P-2E) & 0.84$\pm$0.05 & 0.83$\pm$0.05 & 0.74$\pm$0.07 & 0.78$\pm$0.05 &  -& - \\
 & 3-task (2P-1E) & 0.86$\pm$0.05 & 0.84$\pm$0.05 & 0.75$\pm$0.07 & 0.80$\pm$0.05 &  -& - \\
 & 5-task (1P-1E) & 0.83$\pm$0.05 & 0.82$\pm$0.05 & 0.73$\pm$0.07 & 0.77$\pm$0.05 & 0.87$\pm$0.04 & 0.86$\pm$0.04 \\ \hline
\multirow{3}{*}{\begin{tabular}[c]{@{}l@{}}ctELM on\\ 1.2M\end{tabular}} & 5-task (1P-1E) & 0.86$\pm$0.05 & 0.84$\pm$0.05 & 0.76$\pm$0.07 & 0.80$\pm$0.05 & \textbf{0.88$\pm$0.04} & 0.88$\pm$0.04 \\
 & 5-task (1P-2E) & \textbf{0.87$\pm$0.05} & 0.85$\pm$0.05 & 0.76$\pm$0.07 & \textbf{0.81$\pm$0.05} & \textbf{0.88$\pm$0.04} & \textbf{0.89$\pm$0.03} \\
 & 5-task (2P-1E) & \textbf{0.87$\pm$0.04} & \textbf{0.86$\pm$0.05} & \textbf{0.77$\pm$0.07} & \textbf{0.81$\pm$0.05} & \textbf{0.88$\pm$0.04} & \textbf{0.89$\pm$0.03} \\ \hline
\end{tabular}%
}
\end{table*}

Table~\ref{tab:SC-table} reports the performance analysis of semantic consistency (SC) across model variants and tasks. \textbf{Improved performance over baselines}: Across all tasks, \ctelm{} consistently outperforms all Vec2Text baselines. For example, on abstract reconstruction (\embToAbst{} with penalty=1.2), Vec2Text-sect-ft achieves 0.82, while \ctelm{} models achieve up to 0.87, indicating the effectiveness of the proposed architecture in capturing semantic content from embeddings. \textbf{Impact of repetition penalty}: Applying a repetition penalty of 1.2 yields better semantic consistency than no penalty (1.0),
suggesting that penalizing repetitive token generation encourages more faithful and coherent text generation. \textbf{Effect of task diversity}: Increasing the number of tasks does not degrade performance on individual tasks despite fewer training samples per task. For instance, the SC scores for \embToAbst{} (penalty=1.2) are 0.83 for 1-task, 0.83 for 3-task, and 0.83 for 5-task (all trained with 1P-1E on 190K), indicating that \ctelm{} generalizes well across multitask training regimes without sacrificing single-task quality. \textbf{Effect of data scale}: Scaling the training data from 190K to 1.2M results in consistent improvements across all tasks. For example, \embToSect{} improves from 0.73–0.75 (190K) to 0.76–0.77 (1.2M), and \embToDifferences{} improves from 0.86 to 0.89. This demonstrates the scalability of \ctelm{} and its capacity to benefit from larger datasets. \textbf{Effect of training procedures}: On the smaller 190K dataset, the two-phase training procedure yields superior results. For instance, 3-task (2P-1E) outperforms both 1P-1E and 1P-2E configurations in most tasks. However, on the larger 1.2M dataset, the gap between training procedures narrows. Both 1P-2E and 2P-1E achieve similar top performance (e.g., 0.87 on \embToAbst, 0.81 on \embToPls, 0.89 on \embToDifferences). Notably, 1P-1E requires approximately half the training time and still delivers competitive results, making it a practical alternative for large-scale deployment. We thus use the 5-task 1P-1E model for further experiments.

While Table~\ref{tab:SC-table} evaluates outputs using held-out (but real) test abstracts, Appendix~\ref{sec:interp} investigates \ctelm’s ability to generalize to novel or hypothetical abstracts.
Appendix~\ref{sec:interpolated-generation-examples} presents decoded examples from interplolated embeddings.
For further insight, we analyze \textit{consistency} (versus the original abstract) and \textit{fluency}, both quantitatively---using G-Eval~\cite{liu2023g}--- and qualitatively in Appendix~\ref{sec:geval}.
To test the generalizability of \ctelm{}'s architecture, we also explore the effect of different base chat models in Appendix~\ref{sec:base} and embedding models in Appendix~\ref{sec:emb}.

\section{Validation}

The high Semantic Consistency for clinical trials generated by \ctelm{} from novel embeddings demonstrates that the model has successfully learned a data manifold for a set of abstracts. However, it cannot tell us how well this learned manifold corresponds with a theoretical distribution of parameters of clinical trials. Specifically, it does not tell us (1) whether the abstracts generated from novel embeddings describe clinical trials with likely parameters (in other words, trials that are plausible), or (2) whether directions along the manifold correspond with clinical trial parameters (in other words, whether the geometry of the manifold is clinically meaningful). To further validate \ctelm{} for generative and explanatory use cases, we thus ask two research questions:

\vspace{.5em}
\noindent \textbf{RQ1:} Can \ctelm{} map novel points in the embedding space to plausible clinical trials?
\vspace{.5em}

\noindent \textbf{RQ2:} Are clinical trials generated by \ctelm{} responsive to clinically meaningful directions in the embedding space?
\vspace{.5em}

We seek to answer these questions in the space of abstracts, as, among the tasks \ctelm{} can perform, abstracts most specifically layout the details of one clinical trial. For both questions, we will have \ctelm{} perform the \embToAbst{} task for novel embeddings (those not directly generated from text sources by the mapping $\mathcal{X} \mapsto \mathcal{Z}_{emb}$).

\subsection{Clinical Trial Plausibility}
\label{sec:plaus-hum}
To answer RQ1 (plausibility of clinical trials generated from novel embeddings), we task human experts with discriminating real (test set) abstracts from hypothetical ones generated using the \embToAbst{} prompt and novel embedding vectors. As in \citet{tennenholtzDemystifyingEmbeddingSpaces2024}, we generate novel vectors via interpolation, by averaging randomly selected pairs of test set abstract embeddings. Our main metric is \textit{win rate}, which is the fraction of hypothetical abstracts that successfully fool the expert when paired with a random test abstract. In expectation, the highest achievable win rate is 0.5, meaning generated abstracts are indistinguishable from real ones. As a baseline, we compare to Vec2Text-sect-ft performing the easier task of generating abstracts from original (not interpolated) embeddings, as its section-wise nature precludes direct interpolations. Two experts, both authors, perform the annotation; one an MD (Doctor of Medicine) and one an MBBS (Bachelor of Medicine, Bachelor of Surgery).We randomly select 50 real abstracts from the test set. The first 25 are paired with Vec2Text-sect-ft abstracts for expert 1 and \ctelm{} abstracts for expert 2, while the second 25 are paired with \ctelm{} abstracts for expert 1 and Vec2Text-sect-ft abstracts for expert 2, resulting in 50 single-annotated pairs for each system. Order within pairs is randomized.

\subsection{Concept Activation Vectors}

To answer RQ2 (responsiveness of \ctelm{} outputs to clinically meaningful directions in the embedding space), we follow \citealp{tennenholtzDemystifyingEmbeddingSpaces2024} in moving embeddings along Concept Activation Vectors (CAVs). We train CAVs to identify two axes representing demographics of clinical trial subjects: (1) sex (male vs. female), and (2) age (children vs. older adults). Details of data collection for CAV training can be found in Appendix~\ref{sec:cav-gender}. The model used to identify the CAVs is a linear kernel SVM, implemented with Scikit-learn~\cite{pedregosa2011scikit}. Once CAVs are identified, we add them to embeddings of single sex or single age group clinical trials, with a signed coefficient $\alpha$ determining the strength and direction of modification. The resulting vectors are then normalized to length 1, as other \texttt{BAAI/bge-large-en-v1.5} embeddings, and used to generate new clinical trial abstracts by prompting \ctelm{} to perform the \embToAbst{} task.
To determine responsiveness, we employ an extraction agent to label sex or age of the subjects in the resulting abstracts (see Appendix~\ref{sec:extract}). The complete pipeline is depicted in Figure~\ref{fig:cav-pipeline}.

\subsection{Results}

\begin{figure*}[ht!]
  \includegraphics[width=0.48\linewidth]{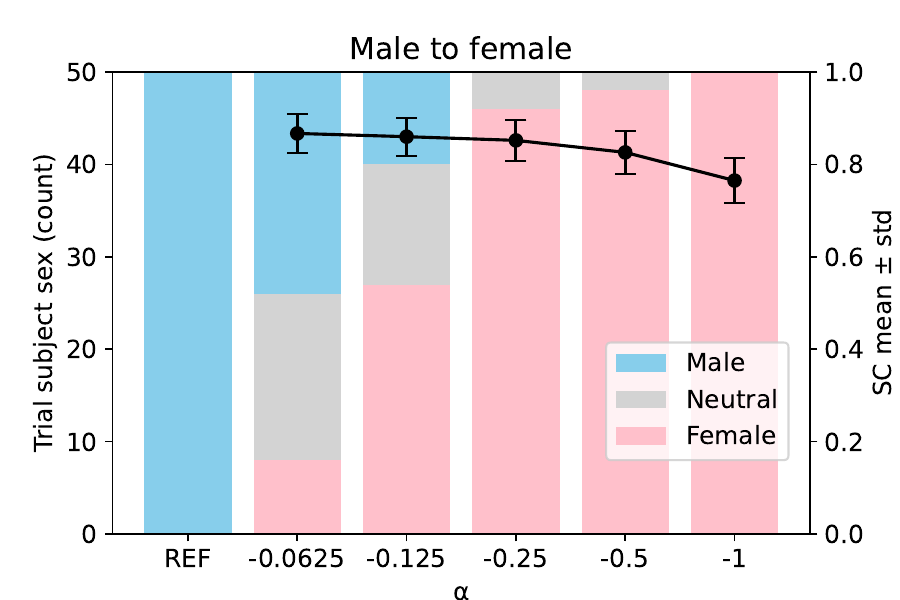} \hfill
  \includegraphics[width=0.48\linewidth]{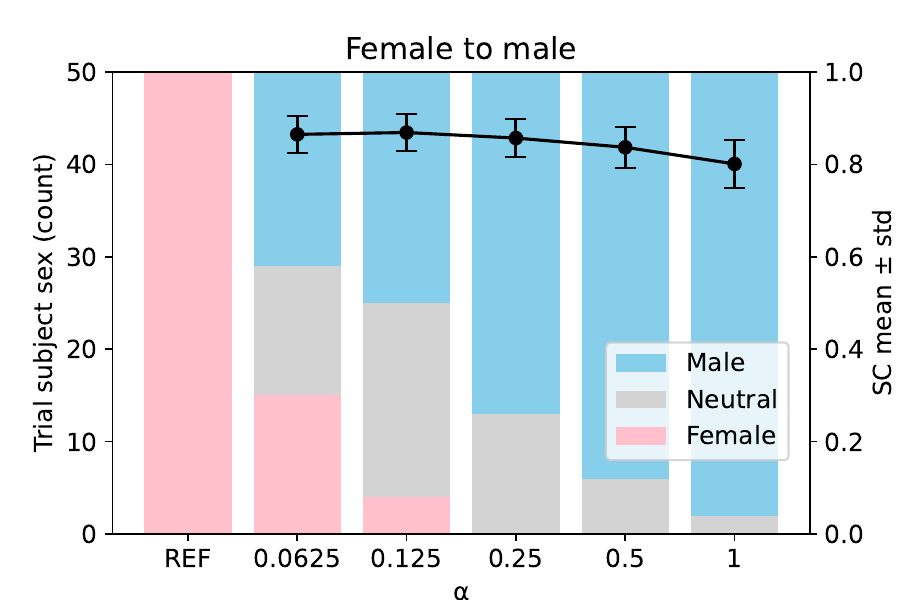}
  \caption {\label{fig:cav-sex}
Moving embeddings along a Concept Activation Vector for sex of trial subjects changes the observed sex in abstracts generated by \ctelm{}. The value $\alpha$ is the coefficient of the added sex vector and thus represents concept strength. Trial subject sex (y-axis, left) refers to the number of trials identified as each sex among a group of 50. REF is sex extracted from original abstracts. Semantic Consistency is shown in black lines (y-axis, right).}
\end{figure*}

\begin{figure*}[ht!]
  \includegraphics[width=0.48\linewidth]{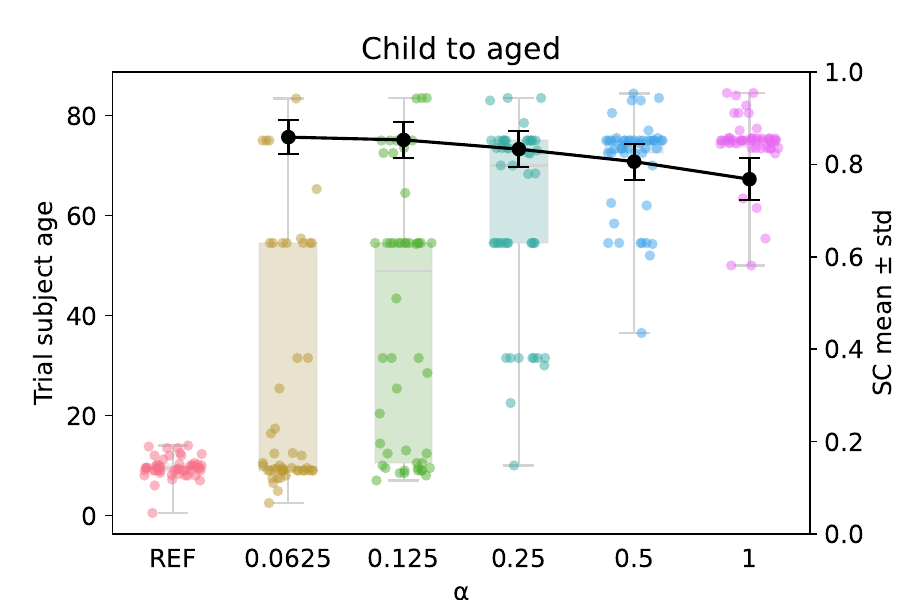} \hfill
  \includegraphics[width=0.48\linewidth]{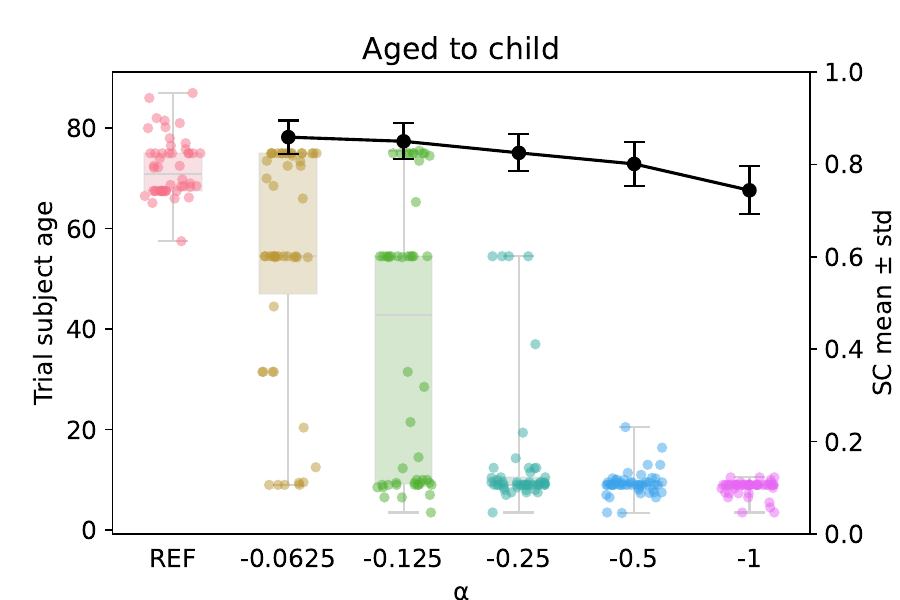}
  \caption {\label{fig:cav-age}Moving embeddings along Concept Activation Vectors for age of trial subjects changes the observed age in abstracts generated by \ctelm{}. The value $\alpha$ is the coefficient of the added age vector. Trial subject age (y-axis, left) refers to the identified age of each trial (each depicted as a point). Box and whisker plots show minima, maxima, medians, and inter-quartile ranges of identified age. Note that horizontal jitter is employed for each discrete $\alpha$ value; the x position of each point within its strip is thus not meaningful. REF is the original abstracts with age extracted directly. Semantic Consistency is shown in black lines (y-axis, right).}
\end{figure*}

\begin{table}[]
\caption{Win rates for human experts}
\label{tab:win-rate-hum}
\resizebox{\columnwidth}{!}{%
\begin{tabular}{@{}llccc@{}}
\hline
&                       & \multicolumn{3}{c}{\textbf{Win rate}}                     \\ \cline{3-5} 

         & \textbf{Embedding}    & \textbf{Exp. 1}        & \textbf{Exp. 2}        & \textbf{Avg.}          \\ \hline\hline
Vec2Text-sect-ft & Original     & 0.00           & 0.04          & 0.02          \\
ctELM    & Interpolated & \textbf{0.48} & \textbf{0.40} & \textbf{0.44} \\
\hline
\end{tabular}}
\end{table}

For RQ1, the win rates of systems vs. human experts are shown in Table~\ref{tab:win-rate-hum}. Vec2Text-sect-ft (the highest performing baseline) fooled experts only once in the 50 pairs, even using unmodified embeddings. On the other hand, \ctelm{} fools the experts 44\% of the time, close to the theoretical limit of 50\%, even though it is performing the more difficult task of generating hypothetical abstracts from novel (interpolated) embeddings.
This shows that \ctelm{} outputs are not only fluent but also describe clinically plausible trials with coherent scientific details. We also develop an automated win rate experiment using an LLM discriminator in order to scale to more conditions (see Appendix~\ref{sec:plaus}).

For RQ2, Figures~\ref{fig:cav-sex} and \ref{fig:cav-age} show results for modification along sex and age CAVs, respectively. Modification successfully changes subject demographics along the expected axes. Both CAVs can even induce intermediate values. For sex, lower values of $\alpha$ produce some abstracts of neutral sex (meaning they include both or do not mention sex), as well as a mixture of male-only and female-only abstracts. For age, lower values of $\alpha$ produce some abstracts with subject ages between children and older adults, as well as abstracts with both extremes. In both cases, semantic consistency remains relatively high as $\alpha$ is increased, though there is some drop-off at full saturation (complete change of subjects), which happens when $\alpha$ is around 1 or -1.



\section{Discussion \& Conclusion}

In this work, we advance the capability of researchers to interpret, explore, and reverse semantic embedding spaces. We show that Embedding Language Models (ELMs), formerly only demonstrated for the domain of film reviews and for proprietary models, generalize to the biomedical domain, specifically for embeddings of clinical trial abstracts, and that lightweight, open-source LLMs can be used as base models for ELMs. We further provide the research community with an open-source architecture and training framework, and we use it to create \ctelm{}, an ELM that can interpret embeddings of clinical trial abstracts. We show that capable ELMs can be trained with very few tasks (1--5 as opposed to 24), and with simpler single-phase training, skipping the adapter pre-training that \citet{tennenholtzDemystifyingEmbeddingSpaces2024} claimed was necessary. Our validation experiments show that, even for novel points with no original text abstract, \ctelm{} can describe clinical trials plausible enough to deceive human experts tasked with discriminating them from real clinical trial abstracts. Our experiments further show that generated abstracts are responsive to changes along clinically meaningful directions in the embedding space. This shows the robustness of the learned mapping and opens many possibilities for language-based interpretation of embedding spaces and controlled generation for diverse synthetic data. 
Taken in total, we expect this work will make aligning LLMs to embedding spaces vastly more accessible, enabling a wide array of downstream applications. We provide our code under the MIT license at \url{https://github.com/BIDS-Xu-Lab/OpenELM}.

\section*{Limitations}

First, we acknowledge that the domain and format of our training data is relatively narrow. As ELMs inherently train to specific data manifolds, it is not clear how well \ctelm{} would generalize to other data from the biomedical domain (such as full articles or clinical notes), let alone data from other domains. For now, we consider ELMs to be corpus- and task-specific (as originally introduced in \citealp{tennenholtzDemystifyingEmbeddingSpaces2024}), and we release our architecture and training code with the hopes that researchers in other domains can easily train bespoke ELMs for other corpora and domains. Future work should test the limits of ELMs for generalizing across domains and tasks.

Second, though we explored fine-tuning Vec2Text, it is still not a perfect baseline since it is based on T5, which has fewer parameters than \ctelm{}'s Llama 3 base model. More in-depth exploration of Vec2Text's iterative correction method of generation, with larger models and domain-specific training data, may improve the viability of this method for exploring embedding spaces and should be explored in the future.

Finally, 
though many generated abstracts were able to deceive human experts when compared to real abstracts, we are still far from translating these into real-world studies. In particular, the applicability of a specific combination of drug, disease, and population would require deep expertise in the relevant field of medicine to validate as a clinical trial. As an example of an ethical hazard, changing of study participants may introduce populations with further protections under Common Rule (such as children or pregnant women) that systems may not account for. Translating our methods and findings into downstream applications will thus require large-scale collaboration with clinicians and bioethicists.



\bibliography{custom}

\appendix

\section{Data Preparation for \embToPls} \label{sec:data-preparation-pls}

\begin{figure}[ht]
    \centering
    \includegraphics[width=\columnwidth]{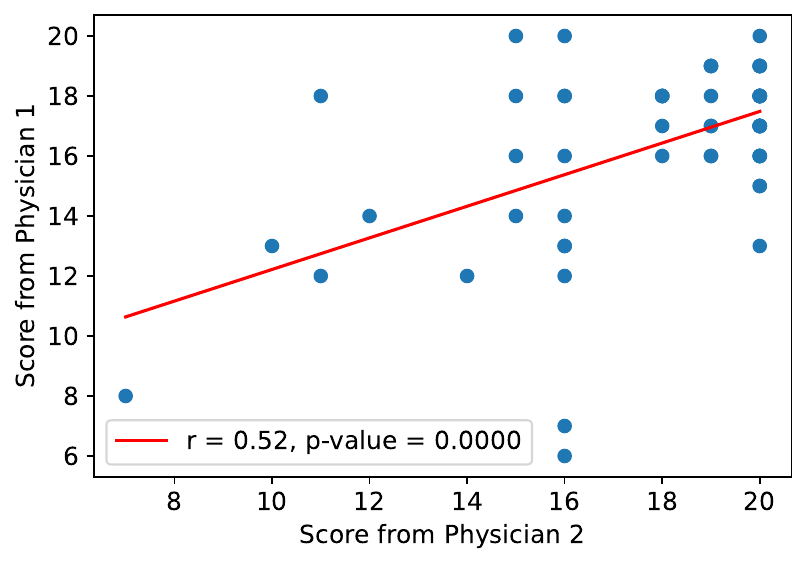}
    \caption{The scatter plot between two physicians' annotated scores with Pearson correlation coefficient ($r$) results.}
    \label{fig:pearson-physicians-scores}
\end{figure}

\begin{figure}
    \centering
    \includegraphics[width=\columnwidth]{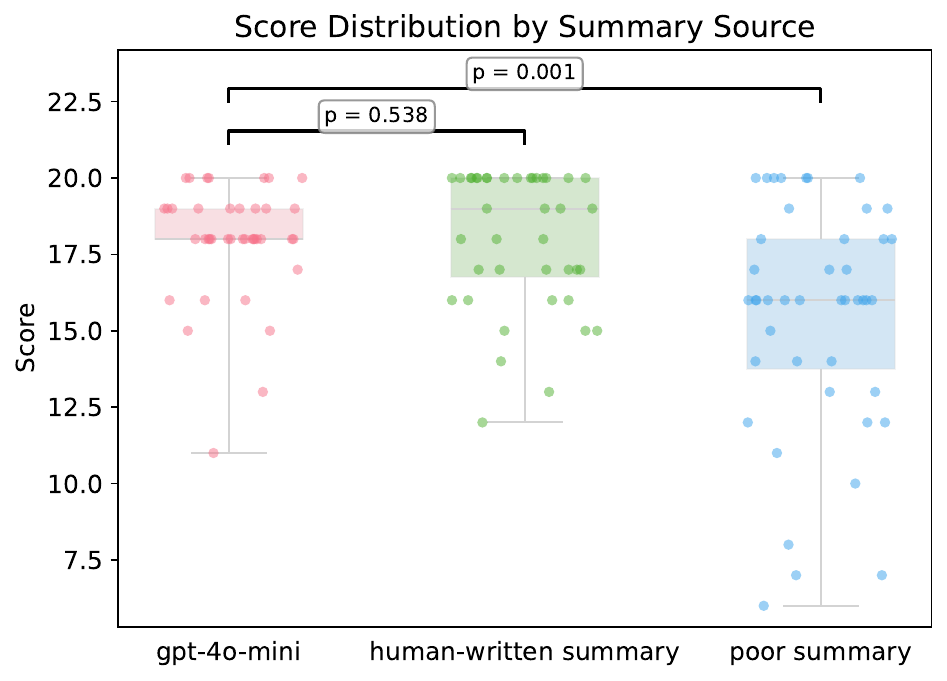}
    \caption{The boxplot for scores by different sources with Wilcoxon rank-sum test results.}
    \label{fig:boxplot-scores-by-sources}
\end{figure}

We generate the plain language summary for each abstract using 
\texttt{gpt-4o-mini} with the prompt shown in Fig.~\ref{fig:pls-prompt}. To measure the quality of the generated summaries, we had two physicians (both women) review 20 sampled summaries generated from \texttt{gpt-4o-mini}. The physicians had a standing contract with our organization to annotate data and had expectations to do similar work and were compensated according to skill level. We discussed with the physicians the goals of the project and role of their evaluations, and defined annotation guidelines (see the attached supplementary file) measuring four aspects: simplicity, accuracy, completeness, and relevance. Each metric is measured using a 5-point Likert scale (1=Poor, and 5=Excellent). To further contextualize these scores, we also have the physicians rate 20 expert-written summaries (which we expect to be of high quality) and 20 summaries of poor quality. Expert and poor-quality summaries were derived from the TREC Plain Language Adaptation of Biomedical Abstracts (PLABA) task (data obtained upon request to organizers). Poor summaries were those with the lowest scores from PLABA's human evaluators. All 60 summaries were shuffled, and their sources were blind to the two physicians. Fig.~\ref{fig:pearson-physicians-scores} presents the scatter plot for scores between two physicians, where the score of each summary is the sum of four metrics. The Pearson correlation coefficient ($r=0.52$) suggests a moderate correlations in two physicians' annotated scores. We conduct the Wilcoxon rank-sum test~\cite{mann-whitney-1947-test} to test whether the median is different between groups (\texttt{gpt-4o-mini} summary versus human-written summary and \texttt{gpt-4o-mini} versus poor summary). Fig.~\ref{fig:boxplot-scores-by-sources} shows that there is no significant difference between median score of \texttt{gpt-4o-mini} summaries and that of human-written summaries. At the same time, there is a significant difference between \texttt{gpt-4o-mini} and the poor summaries, validating the evaluation process. These results suggest the plain language summaries generated by \texttt{gpt-4o-mini} are of sufficient quality for training \ctelm. 

\begin{figure}[t]
\begin{mdframed}
\begin{lstlisting}    
You are a medical writing assistant 
with  expertise in creating plain 
language summaries of scientific 
research. Your goal is to translate
complex scientific abstracts into 
simple, concise summaries 
understandable by a general audience.
Provide only the plain language 
summary, without any additional words, 
instructions, or formatting.

Translate the following PubMed article 
abstract into a plain language summary:

"{abstract}"
\end{lstlisting}
\end{mdframed}
  \caption{The prompt template for generating plain language summary.}
  \label{fig:pls-prompt}
\end{figure}

\section{Data Preparation for \embToCommonalities{} and \embToDifferences} \label{sec:data-preparation-com-dif}

\begin{figure}[t]
    \centering
    \includegraphics[width=\columnwidth]{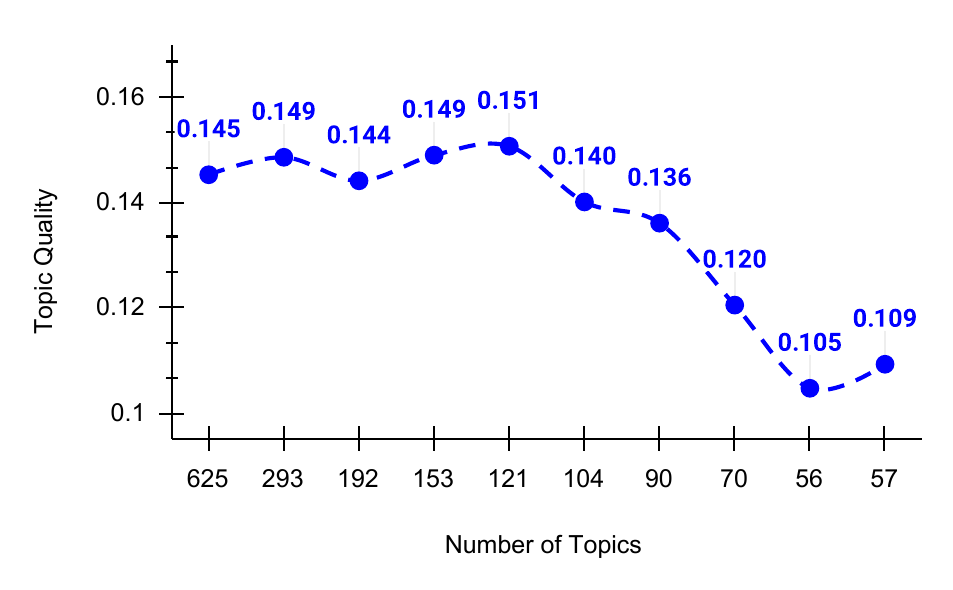}
    \caption{The line plot for helping identify the best number of topics.}
    \label{fig:decide-num-topics}
\end{figure}

We use \texttt{gpt-4o-mini} to generate commonality and difference analyses for each abstract pair, using prompts shown in Fig.~\ref{fig:commonality-prompt} and Fig.~\ref{fig:difference-prompt}. To construct diverse and meaningful abstract pairs, we apply BERTopic~\cite{grootendorst2022bertopic}, a Python-based topic modeling framework, to cluster abstracts in the embedding space. The procedure involves three main steps. First, all abstracts are embedded using \texttt{BAAI/bge-large-en-v1.5} model~\cite{xiao_etal_2024_bge}. Second, we reduce the high-dimensional embeddings into a five-dimensional space using UMAP~\cite{mcinnes2018umap-software} with the following hyperparameters: n\_neighbors=15, n\_components=5, and min\_dist=0.1. Third, HDBSCAN is employed to identify topic clusters within the reduced space. To determine the optimal number of clusters, we search for the min\_cluster\_size value that yields the highest topic quality, measured using the criteria proposed in~\citet{dieng-etal-2020-topic}. As shown in Fig.~\ref{fig:decide-num-topics}, we select min\_cluster\_size=250, resulting in 121 topic clusters with the best topic quality. Using these clusters, we sample abstract pairs either from within the same topic or across different topics. Table~\ref{tab:pair-dist} summarizes the pair distribution used for the \embToCommonalities{} and \embToDifferences{} tasks across training, validation, and testing datasets.

\begin{figure}[t]
\begin{mdframed}
\begin{lstlisting}    
You are an expert in biomedical 
literature analysis.

You are asked to compare two PubMed 
abstracts and identify their 
commonalities. Please use concise 
language. Please directly list five 
commonalities between two abstracts. 
Here are two abstracts:

1. "{abstract1}"
2. "{abstract2}"
\end{lstlisting}
\end{mdframed}
  \caption{The prompt template for listing five commonalities for a abstract pair.}
  \label{fig:commonality-prompt}
\end{figure}

\begin{figure}[t]
\begin{mdframed}
\begin{lstlisting}    
You are an expert in biomedical 
literature analysis.

You are asked to compare two PubMed 
abstracts and identify their 
differences. Please use concise 
language. Please directly list five 
differences between two abstracts. 
Here are two abstracts:

1. "{abstract1}"
2. "{abstract2}"
\end{lstlisting}
\end{mdframed}
  \caption{The prompt template for listing five differences for a abstract pair.}
  \label{fig:difference-prompt}
\end{figure}

\begin{table}[ht]
\caption{Distribution of abstract pairs used for the \embToCommonalities{} and \embToDifferences{} tasks across training, validation, and testing datasets. Pairs are constructed based on topic assignment using BERTopic. Each dataset contains a balanced number of same-topic and different-topic pairs to ensure diversity and control for topic-based variation.}
\label{tab:pair-dist}
\resizebox{\columnwidth}{!}{%
\begin{tabular}{lll}
\hline
\multicolumn{3}{c}{Training Dataset} \\ \hline
 & \embToCommonalities & \embToDifferences \\
Pairs from the Same Topic & 120,897 & 120,897 \\
Pairs from Different Topics & 120,897 & 120,897 \\ \hline
\multicolumn{3}{c}{Validation Dataset} \\ \hline
 & \embToCommonalities & \embToDifferences \\
Pairs from the Same Topic & 1,562 & 1,562 \\
Pairs from Different Topics & 1,564 & 1,564 \\ \hline
\multicolumn{3}{c}{Testing Dataset} \\ \hline
 & \embToCommonalities & \embToDifferences \\
Pairs from the Same Topic & 1,589 & 1,589 \\
Pairs from Different Topics & 1,591 & 1,591 \\ \hline
\end{tabular}%
}
\end{table}

\section{Training Hyperparameters \& Details} \label{sec:training-hyperparameters}

For the first phase of two-phase training procedure, we freeze all model parameters except for embedding adapter $\mathcal{A}$. We then use SFTTrainer in the trl package to optimize the adapter. In the SFTConfig, we set the learning rate as 1e-3, batch size as 4, gradient accumulation steps as 8, and max sequence length as 2,048. With this settings, we train the adapter using AdamW optimizer (default parameters), linear scheduler with a warmup phase, mixed precision (i.e., bfloat16) for one epoch.

As for the second phase of two-phase training procedure and one-phase training procedure, we use SFTTrainer with LoraConfig from peft package. We set the learning rate as 5e-5, batch size as 4, gradient accumulation steps as 8, max grad norm as 1, and max sequence length as 2,048. On the other hand, we set $r$ as 16, lora alpha as 32, lora dropout as 0.05, and bias as none. We only optimze q\_proj and k\_proj in $M_{base}$ and set $\mathcal{A}$ as the module to save. We train the adapter as well as LoRA parameters using AdamW optimizer (default parameters), linear scheduler with a warmup phase, mixed precision (i.e., bfloat16) for one or two epoch(s).

We train the model with the above settings on one Nvidia H100 GPU. The training time for \ctelm{} on 1.2M data using one-phase training procedure for one epoch (1P-1E) takes around 13 hours. On the other hand, the training time for \ctelm{} on 1.2M data using two-phase training procedure for one epoch (2P-1E) takes around 26 hours.

\section{Interpolation Semantic Consistency}
\label{sec:interp}
\begin{table}[ht]
\caption{Performance of semantic consistency on interpolated testing set. \ctelm{} is trained on either 190K or 1.2M data. ($x$P-$y$E) represents $x$-phase training procedure is adopted for $y$ epochs. Best performance for each task is marked in bold.}
\label{tab:sc-interpolation-table}
\resizebox{\columnwidth}{!}{%
\begin{tabular}{lccc}
\hline
Model & \begin{tabular}[c]{@{}l@{}}\embToAbst\\ (pen.=1.2)\end{tabular} & \begin{tabular}[c]{@{}l@{}}\embToAbst\\ (pen.=1.0)\end{tabular} & \embToPls \\ \hline \hline
\multicolumn{4}{c}{\texttt{Llama-3.1-8B-Instruct} (190K training pairs)} \\ \hline
 1-task (1P-1E) & 0.80$\pm$0.03 & 0.79$\pm$0.04 & - \\
 3-task (1P-1E) & 0.80$\pm$0.04 & 0.79$\pm$0.04 & 0.74$\pm$0.04 \\
 3-task (1P-2E) & 0.81$\pm$0.03 & 0.80$\pm$0.04 & 0.75$\pm$0.04 \\
 3-task (2P-1E) & 0.82$\pm$0.03 & 0.81$\pm$0.03 & 0.76$\pm$0.03 \\
 5-task (1P-1E) & 0.80$\pm$0.03 & 0.79$\pm$0.04 & 0.74$\pm$0.04 \\ \hline\hline
\multicolumn{4}{c}{\texttt{Llama-3.1-8B-Instruct} (1.2M training pairs)} \\ \hline
 5-task (1P-1E) & 0.82$\pm$0.03 & 0.81$\pm$0.03 & 0.76$\pm$0.04 \\
 5-task (1P-2E) & \textbf{0.83$\pm$0.03} & 0.81$\pm$0.04 & \textbf{0.77$\pm$0.04} \\
 5-task (2P-1E) & \textbf{0.83$\pm$0.03} & \textbf{0.82$\pm$0.03} & \textbf{0.77$\pm$0.04} \\ \hline\hline
\multicolumn{4}{c}{Gemma 3 (1.2M training pairs, 5-task 1P-1E)} \\ \hline
\texttt{gemma-3-1b-it} & 0.77$\pm$0.04 & - & 0.72$\pm$0.04 \\
\texttt{gemma-3-1b-it} & 0.78$\pm$0.04 & - & 0.73$\pm$0.04 \\
\texttt{medgemma-4b-it} & 0.80$\pm$0.03 & - & 0.74$\pm$0.04 \\
\hline
\end{tabular}%
}
\end{table}

We construct an interpolated testing set by averaging embeddings from randomly selected pairs of test abstracts. This simulates new abstract embeddings that lie between known examples in the semantic space. Across all configurations, \ctelm{} maintains the same observations, such as a repetition penalty of 1.2 leading to better performance than penalty of 1.0 in \embToAbst, and model performance benefiting from more training data. When we compare to the performance with real abstracts (Table~\ref{tab:SC-table}),
although slightly lower, with a drop of roughly 0.02–0.04 points (e.g., $ 0.87 \rightarrow 0.83$ on \embToAbst{}, $0.81 \rightarrow 0.77$ on emb2pls), the scores remain stable and consistent, demonstrating the \ctelm’s robustness in handling unseen, interpolated representations. 

\section{Consistency and Fluency}
\label{sec:geval}

\begin{figure*}
\centering
\small
\begin{tcolorbox}[colback=gray!5, colframe=black, title=Consistency]
\textbf{Criteria:} ``Determine whether the actual output describes the same clinical trial as the input.''

\vspace{1em}

\textbf{Evaluation steps:}
\begin{itemize}
\item ``Check whether the medical condition in `actual output' reflects that of `input'.''
\item ``Check whether the study design (e.g., randomized controlled trial, observational study) in `actual output' reflects that of `input'.'' 
\item ``Check whether the intervention (e.g., drug, therapy) in `actual output' reflects that of `input'.'' 
\item ``Check whether the population (e.g., age group, sex, health status) in `actual output' reflects that of `input'.''
\end{itemize}
\end{tcolorbox}
\caption{Prompts to evaluate \textit{consistency} using G-Eval via the DeepEval framework.}
  \label{fig:consistency}
\end{figure*}

\begin{figure*}
\centering
\small
\begin{tcolorbox}[colback=gray!5, colframe=black, title=Fluency]

\textbf{Criteria}: ``Determine the quality of the 'actual output' in terms of grammar, spelling, punctuation, word choice, and sentence structure.''

\vspace{1em}

\textbf{Evaluation steps:}
\begin{itemize}
\item ``Check whether `actual output' follows standard grammar rules'' 
\item ``Check whether `actual output' is free of spelling and punctuation errors''
\item ``Check whether `actual output' uses appropriate word choice''
\item ``Check whether `actual output' has a coherent sentence structure'' \end{itemize}
\end{tcolorbox}
\caption{Prompts to evaluate \textit{fluency} using G-Eval via the DeepEval framework.}
  \label{fig:fluency}
\end{figure*}

We measure \textit{consistency} and \textit{fluency} quantitatively with G-Eval~\cite{liu2023g}, using the open-source DeepEval framework.\footnote{\url{https://github.com/confident-ai/deepeval}} To define metrics, the framework requires `criteria' and `evaluation steps,' which we provide for consistency and fluency in Figures~\ref{fig:consistency} and~\ref{fig:fluency}, respectively.

Table~\ref{tab:geval} shows G-Eval scores for the best-performing baseline and the \ctelm{} 5-task 1P-1E model. Though consistency of both models can be improved, we find that \ctelm{} has 31\% higher consistency and 317\% higher fluency.

To further investigate these scores, we manually review 25 outputs from Vec2Text-sect-ft and \ctelm{} for errors in consistency and fluency. We analyze these qualitatively by extracting common themes and finding representative examples, as shown in Table~\ref{tab:themes}.

\begin{table*}[]
\centering
\caption{Common types of Consistency and Fluency errors.}
\label{tab:themes}
\begin{tabular}{p{2.5cm}p{3cm}p{8cm}}
\hline\hline
\multicolumn{3}{c}{\textbf{Consistency}}            \\
\hline\hline
\textbf{Model}                    & \textbf{Error type} & \textbf{Examples} \\ \hline
\multirow{5}{*}{ctELM}            & \multirow{4}{*}{\begin{tabular}[c]{@{}l@{}}Imprecision of \\drugs\end{tabular}}                         & “Tropisetron” → “Granisetron” (both 5-HT3 antagonists used as antiemetics) \\ \cline{3-3} &                   & “Telithromycin” → “Clarithromycin” (both antibiotics used to treat pneumonia)     \\ \cline{2-3} 
& \multirow{3}{*}{\begin{tabular}[c]{@{}l@{}}Incorrect patient\\ counts\end{tabular}} & “463 patients” → “1,000 patients” \\ \cline{3-3} &  & “176 eyes of 152 patients” → “100 eyes from 50 patients”     \\ \cline{2-3}  & \multirow{4}{*}{\begin{tabular}[c]{@{}l@{}}Simplification of \\multi-arm studies\end{tabular}}  & Removing a third “minimal contact CB bibliotherapy” group from a study that also included (1) a 6-session Cognitive Behavioral (CB) group and (2) a control group that just got educational brochures.                      \\ \hline
\multirow{16}{*}{Vec2Text-sect-ft} & \begin{tabular}[c]{@{}l@{}}Dropping \\important words\end{tabular} & “daily interruption of sedation” → “daily sedation”    \\ \cline{2-3} & Dropping of statistical results    &   \\ \cline{2-3} & \multirow{5}{*}{\begin{tabular}[c]{@{}l@{}}Jumbling of words\\ and roots\end{tabular}} & “Acupuncture and needle contact were superior to control in reducing the muscle hypertonicity of all muscles except SCM”→"Muscle contact and hypertouch were superior to needle contact in reducing sclerotherapy" \\ \cline{2-3} 
& Numerical imprecision & “age=15.5 years, SD=1.2” → “mean age, 22.5+/-2.5 years”  \\ \cline{2-3} & \multirow{5}{*}{\begin{tabular}[c]{@{}l@{}}Hallucination of\\  nonsensical, \\irrelevant phrases\end{tabular}} & “resected apnea” \\ \cline{3-3} & & “pharmacokinetics of nisoplaban”\\ \cline{3-3} & & “after initial cigarette-dosing”\\ \cline{3-3} & & “a single dose of myosinophils (Meltz, n=30)”\\ \cline{3-3} & & “cumulative femur relapse” \\
\hline\hline
\multicolumn{3}{c}{\textbf{Fluency}}                    \\ \hline\hline
\textbf{Model}  & \textbf{Error type}             & \textbf{Examples}  \\ \hline
\multirow{2}{*}{ctELM}  & Spacing errors      & “Patients'satisfaction”         \\ \cline{2-3}   & Punctuation errors    & “p\textless{}001” (should have a decimal point)  \\ \hline
\multirow{9}{*}{Vec2Text-sect-ft} & \multirow{2}{*}{Incoherent acronyms} & “low-grade tuberculosis (LO)”   \\   &   & “early hip osteoarthritis (EE)”    \\ \cline{2-3}   & \multirow{3}{*}{\begin{tabular}[c]{@{}l@{}}Low-complexity\\ stretches\end{tabular}} & “S-s-s-s-s-s-s-s-s-s-s-s”   \\ \cline{3-3} &   & “100/100/100/100/100 ml”  \\  \cline{3-3}  & & “a single dose of a single dose of a single dose {[}...{]}” \\ \cline{2-3} & \multirow{2}{*}{Illogical phrases}  & “Vitamin D deficiency is a risk factor for developing sun exposure”  \\ \cline{2-3}   & Spelling errors  & “acanemia”  \\ \cline{2-3}  & Punctuation errors   & “P.05” (should have a “\textless{}”)   \\ \hline\hline
\end{tabular}
\end{table*}

\begin{table}[]
\caption{G-Eval consistency and fluency for the best-performing baseline and our model.}
\label{tab:geval}
\resizebox{\columnwidth}{!}{%
\begin{tabular}{lcc}
\hline
Model                  & Consistency & Fluency   \\
\hline\hline
Vec2Text-sect-ft & 0.26$\pm$0.08   & 0.29$\pm$0.12 \\
ctELM (1.2M, 5-task, 1P-1E) & \textbf{0.34$\pm$0.12}   & \textbf{0.92$\pm$0.08} \\
\hline
\end{tabular}}
\end{table}

\section{Effect of Base Chat Model}
\label{sec:base}
In addition to \texttt{Llama-3.1-8B-Instruct}, we also explore 3 variants of Gemma 3~\cite{team2025gemma}, spanning different model sizes and domain-specific training:

\begin{itemize}
    \item  \texttt{gemma-3-1b-it}: 1B parameters, instruction-tuned, text-only, open-domain training.
    \item  \texttt{gemma-3-4b-it}: 4B parameters, instruction-tuned, multi-modal, open-domain training.
    \item  \texttt{medgemma-4b-it}: 4B parameters, multi-modal, instruction-tuned, further pretraining and finetuning on biomedical data starting from \texttt{gemma-3-4b-it} checkpoint.
\end{itemize}

For multimodal models (\texttt{gemma-3-4b-it} and \texttt{medgemma-4b-it}), we load only the text module without the vision module. Results are shown in Table~\ref{tab:base-table}. Performance generally increases with either model size (in parameters) or domain-specific training. Though Llama 3.1 8B performs better than Gemma 3 models, it is not clear if this is due only to number of parameters, as models of equivalent sizes are not available for these two architectures. We also include Gemma 3 models in our analyses in Appendices~\ref{sec:interp} and~\ref{sec:plaus}.

\begin{table*}[htb!]
\caption{Effect of base chat models. Semantic Consistency is shown for 5 tasks on the test set. The \texttt{bge-large-en-v1.5} embedding model is used with the 1P-1E training procedure on the 1.2M 5-task dataset. Mean values over the test set are shown with standard deviations. A repetition penalty of 1.2 is used during inference for \embToAbst. Best performance for each task is marked in bold.}
\label{tab:base-table}
\centering
\begin{tabular}{lccccc}
\hline
Base model & \embToAbst & \embToSect & \embToPls & \embToCommonalities & \embToDifferences \\ \hline \hline
\texttt{Llama-3.1-8B-Instruct} & \textbf{0.86$\pm$0.05} &  \textbf{0.76$\pm$0.07} & \textbf{0.80$\pm$0.05} & \textbf{0.88$\pm$0.04} & \textbf{0.88$\pm$0.04} \\
\texttt{gemma-3-1b-it} & 0.76$\pm$0.05 &  0.69$\pm$0.07 & 0.72$\pm$0.05 & 0.84$\pm$0.04 & 0.79$\pm$0.04 \\
\texttt{gemma-3-4b-it} & 0.79$\pm$0.05 & 0.72$\pm$0.07 & 0.75$\pm$0.05 & 0.86$\pm$0.04 & 0.84$\pm$0.04 \\
\texttt{medgemma-4b-it} & 0.81$\pm$0.05 & 0.73$\pm$0.07 & 0.76$\pm$0.05 & 0.86$\pm$0.04 & 0.85$\pm$0.04 \\ \hline
\end{tabular}
\end{table*}

\section{Effect of Embedding Model}
\label{sec:emb}
To explore how well \ctelm{} generalizes over embedding models, we compare our original \texttt{BAAI/bge-large-en-v1.5} model to two additional options for $E_{emb}$:
\begin{itemize}
    \item \texttt{BAAI/bge-large-en-v1.5} (original): open-domain, 1,024 dimensions, 335M parameters~\cite{xiao_etal_2024_bge}.
    \item \texttt{Alibaba-NLP/gte-large-en-v1.5}: open-domain, 1,024 dimensions, 434M parameters~\cite{zhang2024mgte, li2023towards}.
    \item \texttt{NeuML/pubmedbert-base-embeddings}: biomedical, 768 dimensions, 109M parameters; model of~\citet{gu2021domain} contrastively fine-tuned using Sentence Transformers~\cite{reimers2019sentence}.
\end{itemize}

For all embedding models, we use \texttt{Llama-3.1-8B-Instruct} as the base model and the 1P-1E training procedure on the 1.2M 5-task training set. We find that \ctelm{} generalizes well across embedding models, with \texttt{bge-large-en-v1.5} and \texttt{gte-large-en-v1.5} both exceeding the Vec2Text baselines (Table~\ref{tab:emb}). However, we find here no benefit from the domain-specific \texttt{pubmedbert-base-embeddings} model. This may be due to the smaller number of parameters and fewer embedding dimensions. 

\begin{table*}[]
\caption{Effect of embedding model. Semantic Consistency is shown for 5-task 1P-1E model across embedding models. Mean values over the test set are shown with standard deviations. A repetition penalty of 1.2 is used during inference for \embToAbst. Best performance for each task is marked in bold.}
\label{tab:emb}
\centering
\resizebox{\textwidth}{!}{%
\begin{tabular}{lccccc}
\hline
Embedding model & \embToAbst{}                    & \embToSect{}   & \embToPls{}   & \embToCommonalities{}   & \embToDifferences{}          \\
\hline\hline
\texttt{bge-large-en-v1.5}          & \textbf{0.86$\pm$0.05} & \textbf{0.76$\pm$0.07} & \textbf{0.80$\pm$0.05} & \textbf{0.88$\pm$0.04} & 0.88$\pm$0.04 \\
\texttt{gte-large-en-v1.5}          & 0.83$\pm$0.06 & \textbf{0.76$\pm$0.08} & 0.76$\pm$0.06 & 0.85$\pm$0.04 & \textbf{0.89$\pm$0.03} \\
\texttt{pubmedbert-base-embeddings} & 0.81$\pm$0.09 & 0.65$\pm$0.14 & 0.69$\pm$0.10  & 0.82$\pm$0.07 & 0.83$\pm$0.07 \\
\hline
\end{tabular}}
\end{table*}

\section{Automatic Plausibility Analysis}
\label{sec:plaus}
To measure plausibility of generated clinical trial abstracts under more conditions than feasible with human experts, we develop an LLM-based win rate experiment to mirror the expert version in \S\ref{sec:plaus-hum}.

\subsection{Methods}
As the discriminator agent, we employ \texttt{gpt-4o-2024-11-20} with the prompt shown in Fig.~\ref{fig:prompt-disc}. Though an LLM discriminator may be subject to systematic self-preference~\cite{panickssery2024llm}, using a different model to judge than to generate may ameliorate this problem~\cite{xu2024pride}. To avoid the known phenomenon of positional bias in LLMS~\cite{gu2024survey}, we randomize whether the real abstract is first or second. Further, each win rate for each system is computed 5 times with different random seeds to ensure aggregation over different orderings of each pair.

Additionally, many clinical trial abstracts (real and generated) contain clinical trial registry identifiers. As identifiers may be memorized and associated with study details in the discriminator agent's model, we redacted any such identifiers from both real and generated abstracts. 
We crafted regular expressions to capture each format appearing in World Health Organization's International Clinical Trials Registry Platform\footnote{\raggedright\url{https://www.who.int/tools/clinical-trials-registry-platform}} as of February 10, 2025  (Table~\ref{tab:regex}) and replaced matches with ``\texttt{[redacted]}''. 

\begin{figure}[t]
\begin{mdframed}
\begin{lstlisting}    
Which of the following abstracts is
more likely to be a real abstract
describing a clinical trial? Return
only "1" or "2".

1. "{abstract1}"

2. "{abstract2}"
\end{lstlisting}
\end{mdframed}
  \caption{The discriminator agent prompt. Order of real and generated abstracts is randomized.}
  \label{fig:prompt-disc}
\end{figure}

\begin{table*}[ht!]
\centering
{
\begin{tabular}{ll}
\textbf{Regex} & \textbf{Registry} \\
\hline
\texttt{ACTRN[0-9]+} & Australian New Zealand Clinical Trials Registry \\
\texttt{ChiCTR[A-Z0-9-]+} & Chinese Clinical Trials Register \\
\texttt{CTIS[0-9-]+} & European Union Clinical Trials Information System \\
\texttt{CTRI[0-9/]+} & Clinical Trials Registry - India \\
\texttt{DRKS[0-9]+} & German Clinical Trials Register \\
\texttt{EUCTR[0-9a-zA-Z-]+} & European Clinical Trials Register \\
\texttt{IRCT[0-9]+N[0-9]+} & Iranian Registry of Clinical Trials \\
\texttt{ISRCTN[0-9]+} & UK Clinical Study Register \\
\texttt{ITMCTR[0-9]+} & International Traditional Medicine Clinical Trial Registry \\
\texttt{JPRN-[a-zA-Z0-9]+} & Japan Primary Registries Network \\
\texttt{KCT[0-9]\{7\}} & Korean Clinical Research Information Service \\
\texttt{LBCTR[0-9]+} & Lebanese Clinical Trials Registry \\
\texttt{NCT[0-9]\{8\}} & US National Clinical Trial \\
\texttt{NL-OMON[0-9]+} & Overview of Medical Research in the Netherlands \\
\texttt{PACTR[0-9]+} & Pan African Clinical Trials Registry \\
\texttt{RBR-[a-z0-9]+} & Brazilian Clinical Trials Registry \\
\texttt{RPCEC[0-9]\{4\}} & Cuban Registry of Clinical Trials \\
\texttt{SLCTR/\textbackslash d+/\textbackslash d+} & Sri Lanka Clinical Trials Registry \\
\texttt{TCTR[0-9]+} & Thai Clinical Trials Registry \\
\end{tabular}}
\caption{Regular Expressions for identifying Clinical Trial Registry identifiers.}
\label{tab:regex}
\end{table*}

\subsection{Results}
The LLM-based win rates of abstracts from interpolated embeddings and from embeddings moved along CAVs are shown in Table~\ref{tab:win-rate}. First, we note the overall similarity in the automatic results to human results for the conditions tested by both. Specifically, Vec2Text-sect-ft (with original embeddings) achieves a win rate of 0.02 vs. experts and 0.01 vs. the LLM, whereas \ctelm{} based on \texttt{Llama-3.1-8B-Instruct} (with interpolated embeddings) achieves a win rate of 0.44 vs. experts and 0.40 vs. the LLM. LLM-based win rates are in fact slightly lower than their human counterparts, meaning they were better able to discriminate. This suggests the LLM discriminator agent is reliable enough to provide useful results for other conditions.

Second, we note that using novel embeddings has little on affect plausibility of abstracts generated by \ctelm{}. For \texttt{Llama-3.1-8B-Instruct}, interpolated embeddings and those produced using age CAVs in fact have slightly higher win rates than for original embeddings. Though there are drops for interpolated vs. original embeddings for Gemma 3 models, their inter-quartile ranges still overlap. These results further suggests that \ctelm{} has learned not only to create fluent text descriptions of existing clinical trials, but has learned a manifold of plausible clinical trials.

Finally, we note that win rates are lower for Gemma 3 models. This is not unexpected, given their lower Semantic Consistency. It is also in line with expectations that the large 4B parameter model performs better than the 1B parameter model. However, we see no benefit here of the continued domain-based pretraining and fine-tuning of \texttt{medgemma-4b-it}, which is suprising. It is possible that the narrow focus of this models additional training tasks (mostly multiple choice questions) affected their general language modeling or instruction following capabilities without benefiting this particular biomedical task.

\begin{table*}[htb!]
\caption{Win Rate of abstracts generated from original or modified embedding vectors when presented to an LLM discriminator along with a real abstract. For \ctelm{}, the 1P-1E training procedure is used with the 5-task 1.2M dataset, and all CAVs are applied with $|\alpha|=0.5$.}
\centering
\label{tab:win-rate}
\begin{tabular}{llcccc}
\hline
&                       & \multicolumn{4}{c}{\textbf{Win rate by embedding type}}                     \\ \cline{3-6} 
\textbf{Method}   & \textbf{Base model}      & \textbf{Orig.} & \textbf{Interp.}   & \textbf{CAV-sex} & \textbf{CAV-age}   \\
\hline\hline
Vec2Text          & -     & 0.00±0.00       &         -          &      -           &   -                 \\
Vec2Text-ft     & -       & 0.01±0.00       &        -           &       -          &   -                 \\
Vec2Text-sect     & -     & 0.00±0.00       &        -            &       -          &    -                \\
Vec2Text-sect-ft  & -    & 0.01±0.00       &           -         &          -        &    -                \\
ctELM & \texttt{gemma-3-1b-it}  & 0.12±0.02      & 0.13±0.05          &   -             &  -                  \\
ctELM & \texttt{gemma-3-4b-it}  & 0.31±0.07      & 0.29±0.06          &    -            &   -                 \\
ctELM & \texttt{medgemma-4b-it} & 0.22±0.05      & 0.16±0.02          &       -           &  -                  \\
ctELM & \texttt{Llama-3.1-8B-Instruct} & \textbf{0.39±0.06}      & \textbf{0.40±0.06} & \textbf{0.38±0.07}        & \textbf{0.40±0.08}
\\ \hline
\end{tabular}
\end{table*}

\section{Data Collection for Sex Concept Activation Vector}
\label{sec:cav-gender}

\begin{figure*}[t]
\centering
  \includegraphics[width=\textwidth]{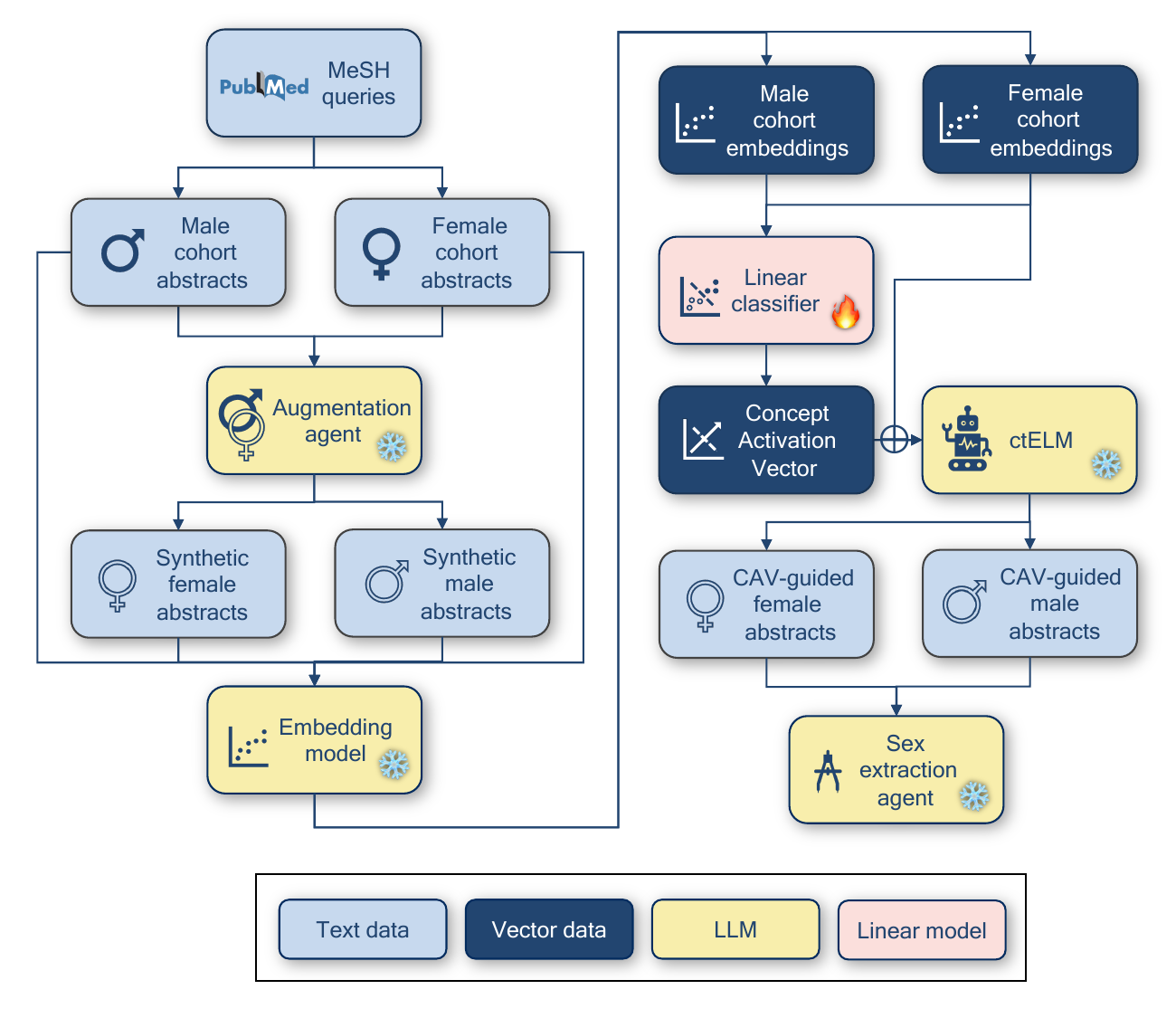}
  \caption{The workflow for modifying clinical trial embeddings with Concept Activation Vectors, including data collection, augmentation, linear model training, generation from modified embeddings, and evaluation. Depicted here is the sex CAV; the same workflow applies to age, except with child and aged instead of male and female.}
  \label{fig:cav-pipeline}
\end{figure*}

To identify a symmetric axis of cohort sex, we collected clinical trials describing interventions that \textit{could} apply to both sexes, but that happened to only include one sex in the study cohort. We thus searched PubMed for randomized controlled trials with only one of the MeSH terms `Male' and `Female,' and excluding studies with MeSH terms related to sex-specific conditions (such as prostate cancer) or procedures (such as hysterectomy). We further required one of four gendered nouns (`men,' `women,' `boys,' `girls') to appearing in the `Title/Abstract field' to filter out studies with single-sex MeSH terms but no mention of cohort sex in abstracts. The complete PubMed search strings are provided in Figs.~\ref{fig:mesh-male} and \ref{fig:mesh-female}. Search results were sorted using the `Best Match' option, and, for each sex, the first 25 were collected that satisfied two manually verified criteria: (1) cohort sex was mentioned in the abstract (not just the title), and (2) there were no implied participants of the opposite sex, for example, ``Study participants: 50 adults (23 women; 46\%).'' We then augmented the data to ensure semantically symmetrical pairs by using \texttt{gpt-4o-2024-11-20} (depicted as `Augmentation agent' in Fig.~\ref{fig:cav-pipeline}) to reverse the sex of study participants in these initial 50 abstracts, using the prompt in Fig.~\ref{fig:sex-prompt}. This created a total of 100 abstracts describing clinical trials: 25 real male, 25 real female, 25 synthetic male, and 25 synthetic female. All synthetic samples were manually reviewed for successful change of cohort sex and consistency of other study details.

\begin{figure*}[t]
\begin{mdframed}
\begin{lstlisting}    
(
  English[Language]
  AND (Randomized Controlled Trial [Publication Type])
   
  AND (Male[MeSH Terms])
  NOT (Female[MeSH Terms])
  
  NOT (Genitalia[MeSH Terms])
  NOT (Urogenital Diseases[MeSH Terms])
  NOT (Pelvic Neoplasms[MeSH Terms])
  NOT (Urogenital Surgical Procedures[MeSH Terms])
  NOT (Fertility Preservation[MeSH Terms])
  NOT (Contraceptive Devices[MeSH Terms])
  NOT (Alopecia[MeSH Terms])
  NOT (Gonadal Disorders[MeSH Terms])
  NOT (Gonadal Hormones[MeSH Terms])
) AND
(
  ("2024/01/01"[EPDAT] : "2024/12/31"[EPDAT])
) AND
(
  (men[Title/Abstract]) OR (boys[Title/Abstract])
) 
\end{lstlisting}
\end{mdframed}
  \caption{The PubMed search string for male single-sex clinical trials.}
  \label{fig:mesh-male}
\end{figure*}

\begin{figure*}[t]
\begin{mdframed}
\begin{lstlisting}    
(
  English[Language]
  AND (Randomized Controlled Trial[Publication Type])
    
  AND (Female[MeSH Terms])
  NOT (Male[MeSH Terms])
  
  NOT (Pregnancy[MeSH Terms])
  NOT (Menopause[MeSH Terms])
  NOT (Genitalia[MeSH Terms])
  NOT (Urogenital Diseases[MeSH Terms])
  NOT (Breast Neoplasms[MeSH Terms])
  NOT (Pelvic Neoplasms[MeSH Terms])
  NOT (Urogenital Surgical Procedures[MeSH Terms])
  NOT (Menstruation Disturbances[MeSH Terms])
  NOT (Osteoporosis, Postmenopausal[MeSH Terms])
  NOT (Fertility Preservation[MeSH Terms])
  NOT (Contraceptive Devices[MeSH Terms])
  NOT (Gonadal Disorders[MeSH Terms])
  NOT (Gonadal Hormones[MeSH Terms])
) AND
(
  ("2024/01/01"[EPDAT] : "2024/12/31"[EPDAT])
) AND
(
  (women[Title/Abstract]) OR (girls[Title/Abstract])
) 
\end{lstlisting}
\end{mdframed}
  \caption{The PubMed search string for female single-sex clinical trials.}
  \label{fig:mesh-female}
\end{figure*}

\begin{figure*}[t]
\begin{mdframed}
\begin{lstlisting}    
Modify this abstract so the subjects are {'male','female'} rather than
{'female','male'}. Output only the abstract, with no quotes or formatting.

"{abstract}"
\end{lstlisting}
\end{mdframed}
  \caption{The prompt template for symmetric augmentation of abstracts for the sex Concept Activation Vector.}
  \label{fig:sex-prompt}
\end{figure*}

\section{Data Collection for Age Concept Activation Vector}
\label{sec:cav-gender}

Similarly to the sex Concept Activation Vector, we collected clinical trials describing interventions that \textit{could} apply to both children and aged subjects, but that happened to only include one of these groups in the trial. We thus searched PubMed for randomized controlled trials with only one of the top-level MeSH age groups `Child' (defined in MeSH as ages 6-12) and `Aged' (defined as age 65 and above), further excluding the other top-level groups (`Infant', `Child, Preschool', `Adolescent', `Adult', and `Middle Aged'). To find studies applicable across ages, we also excluded age-specific study elements, such as `Schools' for child studies or `Dementia' for aged studies. The complete PubMed search strings are provided in Figs.~\ref{fig:mesh-child} and \ref{fig:mesh-aged}. Again, search results were sorted using the `Best Match' option, and, for each age group, the first 25 were collected that satisfied two manually verified criteria: (1) subject age was mentioned in the abstract, and (2) there were no implied participants of other ages. We again augmented the data to ensure semantically symmetrical pairs by using \texttt{gpt-4o-2024-11-20} to reverse the age of study participants in these initial 50 abstracts, using the prompt in Fig.~\ref{fig:age-prompt}. This created a total of 100 abstracts describing clinical trials: 25 real child, 25 real aged, 25 synthetic child, and 25 synthetic aged. All synthetic samples were manually reviewed for successful change of subject age and consistency of other study details.

\begin{figure*}[t]
\begin{mdframed}
\begin{lstlisting}    
(
  English[Language]
  AND (Randomized Controlled Trial[Publication Type])
   
  AND (Child[MeSH Terms])
  NOT (Child, Preschool[MeSH Terms])
  NOT (Infant[MeSH Terms])
  NOT (Adolescent[MeSH Terms])
  NOT (Adult[MeSH Terms])
  NOT (Middle Aged[MeSH Terms])
  NOT (Aged[MeSH Terms])
  
  NOT (Immunization Schedule[MeSH Terms])
  NOT (Child Behavior[MeSH Terms)
  NOT (Growth Disorders[MeSH Terms])
  NOT (Growth Hormone[MeSH Terms])
  NOT (Growth and Development[MeSH Terms])
  NOT (Tooth, Deciduous[MeSH Terms])
  NOT (Child Abuse[MeSH Terms])
  NOT (Family[MeSH Terms])
  NOT (Schools[MeSH Terms])
  NOT (Curriculum[MeSH Terms])
  NOT (Congenital, Hereditary, and Neonatal Diseases and Abnormalities
  [MeSH Terms])
  NOT (Neurodevelopmental Disorders[MeSH Terms])
) AND
(
  ("2024/01/01"[EPDAT] : "2024/12/31"[EPDAT])
)
\end{lstlisting}
\end{mdframed}
  \caption{The PubMed search string for child single-age-group clinical trials.}
  \label{fig:mesh-child}
\end{figure*}

\begin{figure*}[t]
\begin{mdframed}
\begin{lstlisting}    
(
  English[Language]
  AND (Randomized Controlled Trial[Publication Type])
   
  AND (Aged[MeSH Terms])
  NOT (Child[MeSH Terms])
  NOT (Child, Preschool[MeSH Terms])
  NOT (Infant[MeSH Terms])
  NOT (Adolescent[MeSH Terms])
  NOT (Middle Aged[MeSH Terms])
  
  NOT (Breast Neoplasms[MeSH Terms])
  NOT (Dementia[MeSH Terms])
  NOT (Polypharmacy[MeSH Terms])
  NOT (Activities of Daily Living[MeSH Terms])
) AND
(
  ("2024/01/01"[EPDAT] : "2024/12/31"[EPDAT])
)
\end{lstlisting}
\end{mdframed}
  \caption{The PubMed search string for aged single-age-group clinical trials.}
  \label{fig:mesh-aged}
\end{figure*}

\begin{figure*}[t]
\begin{mdframed}
\begin{lstlisting}    
Modify this abstract so the subjects are {'children','older adults'} rather
than {'older adults','children'}. Include specific ages. Output only the 
abstract, with no quotes or formatting.

"{abstract}"
\end{lstlisting}
\end{mdframed}
  \caption{The prompt template for symmetric augmentation of abstracts for the age Concept Activation Vector.}
  \label{fig:age-prompt}
\end{figure*}

\section{Extraction of Subject Demographics}
\label{sec:extract}
To evaluate responsiveness of \ctelm{} to CAVs, we employed an extraction agent comprising \texttt{gpt-4o-2024-11-20} with the system messages depicted in Figures~\ref{fig:prompt-get-sex} and \ref{fig:prompt-get-age} for sex and age, respectively. For both, the user prompt template was ``Now process the following abstract: \{abstract\}".

\begin{figure*}[t]
\begin{mdframed}
\begin{lstlisting}    
You are a biomedical natural language processing assistant. Given the abstract
of a clinical trial study, your task is to identify the gender of the study
population.

Your output must be in the following JSON format:
{
  "gender": "female"  // or "male" or "neutral"
}

Guidelines:
- If the abstract mentions that the study participants are women or females,
output "female".
- If the abstract mentions men or males, output "male".
- If the abstract only mentions the number of participants without specifying
gender, output "neutral".
- If both male and female participants are mentioned and the study includes
both, still output "neutral".
- Do not infer gender based on disease or context. Only use explicit statements.
\end{lstlisting}
\end{mdframed}
  \caption{The subject sex extraction agent system message.}
  \label{fig:prompt-get-sex}
\end{figure*}

\begin{figure*}[t]
\begin{mdframed}
\begin{lstlisting}    
You are a biomedical natural language processing assistant. Given the abstract
of a clinical trial study, your task is to extract or infer the average age (in
years) of the study population.

Your output must be in the following JSON format:
{
  "age": 54.3  // numerical value only
}

Guidelines:
1. If the study mentions the **mean or average age**, extract and return that
value.
2. If the study mentions an **age range** (e.g., "30 to 50 years"), compute the
average (e.g., (30+50)/2 = 40.0) and return that value.
3. If no explicit age value is mentioned, infer the most likely average age
based on population group terms in the text, using this mapping:

- "Child, Preschool": 2-5 years  -> 3.5
- "Child": 6-12 years  -> 9
- "Adolescent": 13-18 years  -> 15.5
- "Adult": 19-44 years  -> 31.5
- "Middle Aged": 45-64 years  -> 54.5
- "Aged": 65+ years  -> 75
- "Aged, 80 and over": 80+ years  -> 85
- "Octogenarians": 80-89 years  -> 84.5
- "Nonagenarians": 90-99 years  -> 94.5
- "Centenarians": 100+ years  -> 100

Choose the most appropriate inferred value if only a population label is
present.

Only include the JSON output. Do not explain or add commentary.
\end{lstlisting}
\end{mdframed}
  \caption{The subject age extraction agent system message.}
  \label{fig:prompt-get-age}
\end{figure*}

\section{Generated Examples of Interpolated Embedding} \label{sec:interpolated-generation-examples}

To illustrate the generative capabilities of \ctelm{} on interpolated embeddings, we present three examples of generated texts (background, objective, and result) derived from the average of two distinct abstract embeddings. As shown in Fig.~\ref{fig:example-generated-background}, Fig.~\ref{fig:example-generated-objective}, and Fig.~\ref{fig:example-generated-result}, \ctelm{} successfully synthesizes semantically coherent sentences that reflect thematic overlaps between the paired source abstracts. This demonstrates the model’s capacity to interpolate meaningfully between known research areas. Importantly, this capability suggests a promising avenue for exploratory scientific hypothesis generation. For instance, by sampling embeddings from underrepresented or ``empty'' regions of the semantic space (i.e., areas not directly covered by existing training data), \ctelm{} could be prompted to generate novel study hypotheses, bridging previously unconnected biomedical concepts. This highlights the potential of embedding-based generation as a tool for ideation and discovery in literature-based research.

\begin{figure*}
\centering
\begin{tcolorbox}[colback=gray!5, colframe=black, title=Example 1: Generated Background of Interpolated Embedding]
\textbf{Generated Background Section of Interpolated Embedding between PMID=24099432 and PMID=17064200}: 
``This paper presents baseline data from a randomized clinical trial examining the effectiveness of a \emph{cognitive behavioral intervention (CBI)} for improving medication adherence and depression outcomes among \emph{patients with poorly controlled hypertension}.''

\vspace{1em}

\textbf{Partial Abstract for PMID=24099432 (Cognitive Behavioral Therapy, Depressive disorder)}: ``[Objective] We tested whether a brief cognitive behavioral (CB) group and bibliotherapy prevention reduce major depressive disorder (MDD) onset, depressive symptoms, and secondary outcomes relative to brochure controls in adolescents with self-reported depressive symptoms when school personnel recruit participants and deliver the intervention. … [Results] The finding that a brief CB group intervention delivered by real-world providers significantly reduced MDD onset relative to both brochure control and bibliotherapy is very encouraging, although effects on continuous outcome measures were small or nonsignificant and approximately half the magnitude of those found in efficacy research, potentially because the present sample reported lower initial depression.''

\vspace{1em}

\textbf{Partial Abstract for PMID=17064200 (Hypertension)}: ``[Objective] To examine potential threats to internal and external study validity caused by differential patient withdrawal from a randomized controlled trial evaluating pharmacist management of hypertension, to compare the characteristics of patients who withdrew with those of patients who completed the study, and to identify characteristics that predispose patients to withdraw from hypertension management. … [Results] Therefore, internal validity was preserved, and outcomes from the study groups could be reliably compared. A lack of significant differences between patients who withdrew versus those who completed, with the exception of insurance status, suggests that external validity was not jeopardized.''
\end{tcolorbox}
\caption{Example of generated background from interpolated embedding between two clinical trials. The generated text reflects a synthesis of themes related to cognitive behavioral therapy and hypertension.}
  \label{fig:example-generated-background}
\end{figure*}

\begin{figure*}
\begin{tcolorbox}[colback=gray!5, colframe=black, title=Example 2: Generated Objective Section of Interpolated Embedding]
\textbf{Generated Objective of Interpolated Embedding between PMID=15914575 and PMID=9777179}: 
``A study was conducted to determine if irritable bowel syndrome (IBS) patients with depressive symptoms have a better response to a selective serotonin reuptake inhibitor (SSRI) than those without depressive symptoms.''

\vspace{1em}

\textbf{Partial Abstract for PMID=15914575 (Irritable bowel syndrome)}: ``[Background] Melatonin, a sleep promoting agent, is involved in the regulation of gastrointestinal motility and sensation. [Objective] We aimed to determine if melatonin was effective in improving bowel symptoms and sleep disturbances in irritable bowel syndrome (IBS) patients with sleep disturbance. ... [Results] The findings suggest that the beneficial effects of melatonin on abdominal pain in IBS patients with sleep disturbances are independent of its action on sleep disturbances or psychological profiles.''

\vspace{1em}

\textbf{Partial Abstract for PMID=9777179 (Chronic mental illness)}: ``[Objective] We sought to identify baseline predictors of response to clozapine. Data were from a 15-site randomized clinical trial comparing clozapine and haloperidol in hospitalized patients with refractory schizophrenia (n = 423). Three-month outcomes were analyzed with the full sample (n=368 due to attrition). ... [Results] Although high levels of symptoms were associated with greater improvement on clozapine, these findings are not robust enough to suggest that any specific, clinically defined subgroup of refractory patients should be preferentially targeted for clozapine treatment.''
\end{tcolorbox}
\caption{Generated objective from the interpolation of abstract embeddings concerning irritable bowel syndrome and chronic mental illness. The output proposes a novel trial scenario integrating elements of both parent studies.}
  \label{fig:example-generated-objective}
\end{figure*}

\begin{figure*}
\begin{tcolorbox}[colback=gray!5, colframe=black, title=Example 3: Generated Results Section of Interpolated Embedding]
\textbf{Generated Result of Interpolated Embedding between PMID=15166570 and PMID=12860586}: 
``The results of this study demonstrate that patients with functional constipation have a better outcome after treatment with a single dose of subcutaneous diamorphine than after treatment with a single dose of subcutaneous hyoscine butylbromide.''

\vspace{1em}

\textbf{Partial Abstract for PMID=15166570 (Opioids, Morphine, Urinary tract dysfunction)}: ``[Background] Intrathecal administration of opioids may cause lower urinary tract dysfunction. In this study, the authors compared the effects of morphine and sufentanil administered intrathecally in a randomized double-blind fashion (two doses each) on lower urinary tract function in healthy male volunteers. ... [Conclusion] Intrathecal opioids decrease bladder function by causing dose-dependent suppression of detrusor contractility and decreased sensation of urge. Recovery of normal lower urinary tract function is significantly faster after intrathecal sufentanil than after morphine, and the recovery time is clearly dose dependent.''

\vspace{1em}

\textbf{Partial Abstract for PMID=12860586 (Dyspepsia)}: ``[Background] The value of the test-and-treat strategy in the approach to dyspepsia has been evaluated only in a few secondary care studies. Most patients with dyspepsia, however, are treated by their primary care physician ... [Conclusion] The test-and-treat strategy proved to be as effective and safe as prompt endoscopy. Only a minority of patients were referred for endoscopy after the test-and-treat approach.''
\end{tcolorbox}
\caption{Example of a generated result sentence from interpolated embeddings of abstracts on opioids and dyspepsia. The output blends insights into drug response and gastrointestinal outcomes, demonstrating semantic consistency across domains.}
  \label{fig:example-generated-result}
\end{figure*}

\end{document}